\documentclass{nature}

\usepackage[super, numbers]{natbib}

\usepackage[utf8]{inputenc}
\usepackage{graphicx}
\usepackage{multirow}
\usepackage{enumitem}
\usepackage{color}
\usepackage{subcaption}
\usepackage{amsmath}
\usepackage{amsfonts}
\usepackage{siunitx}
\usepackage{footmisc}
\usepackage{hyperref}

\renewcommand{\cite}[1]{ \citep{#1}}
\usepackage[normalem]{ulem}

\usepackage{xargs}

\makeatletter
\def\blfootnote{\xdef\@thefnmark{}\@footnotetext}
\makeatother

\title{Unsupervised Behaviour Analysis and Magnification (uBAM) using Deep Learning}

\author{Biagio Brattoli$^{1\star\dagger}$,
Uta B\"uchler$^{1\star}$,
Michael Dorkenwald$^{1}$,
Philipp Reiser$^{1}$,
Linard Filli$^{2,3}$,
Fritjof Helmchen$^{4,5}$,
Anna-Sophia Wahl$^{4,5}$,
Bj\"orn Ommer$^{1\dagger}$}

\begin{document}
{\centering\Large\textbf{Unsupervised Behaviour Analysis and Magnification (uBAM) \mbox{using} Deep Learning}}

\noindent\textbf{Authors: }
Biagio Brattoli$^{1\star}$, 
Uta B\"uchler$^{1\star}$,
Michael Dorkenwald$^{1}$,
Philipp Reiser$^{1}$,
Linard Filli$^{2,3}$,
Fritjof Helmchen$^{4,5}$,
Anna-Sophia Wahl$^{4,5,6\star}$,
Bj\"orn Ommer$^{1\star\dagger}$

\noindent$\star$ Equal first and last authorship

\noindent\textbf{Affiliations: }\\
\noindent$^1$Interdisciplinary Center for Scientific Computing \& Heidelberg Collaboratory for Image Processing, Heidelberg University, Germany.\\
$^2$Department of Neurology, University Hospital and University of Zurich, Zurich, Switzerland. \\
$^3$Spinal Cord Injury Center, Balgrist University Hospital, Zurich, Switzerland. \\
$^4$Brain Research Institute, University of Zurich, Zurich, Switzerland.\\
$^5$Neuroscience Center Zurich, Zurich, Switzerland.\\
$^6$Central Institute of Mental Health, Heidelberg University, Mannheim, Germany

\noindent $\dagger$ Correspondence to: \\
Bj\"orn Ommer (ommer@uni-heidelberg.de)

\noindent \textbf{Official article:} At \textit{Nature Machine Intelligence} \textcolor{blue}{\url{https://rdcu.be/ch6pL}}

\clearpage

\begin{abstract}

\vspace{0.5cm}
\section*{SUMMARY}
Motor behaviour analysis is essential to biomedical research and clinical diagnostics as it provides a non-invasive strategy for identifying motor impairment and its change caused by interventions. State-of-the-art instrumented movement analysis is time- and cost-intensive, since it requires placing physical or virtual markers. Besides the effort required for marking keypoints or annotations necessary for training or finetuning a detector, users need to know the interesting behaviour beforehand to provide meaningful keypoints.
We introduce unsupervised behaviour analysis and magnification (uBAM), an automatic deep learning algorithm for analysing behaviour by discovering and magnifying deviations. 
A central aspect is unsupervised learning of posture and behaviour representations to enable an objective comparison of movement. Besides discovering and quantifying deviations in behaviour, we also propose a generative model for visually magnifying subtle behaviour differences directly in a video without requiring a detour via keypoints or annotations. Essential for this magnification of deviations even across different individuals is a disentangling of appearance and behaviour. Evaluations on rodents and human patients with neurological diseases demonstrate the wide applicability of our approach. Moreover, combining optogenetic stimulation with our unsupervised behaviour analysis shows its suitability as a non-invasive diagnostic tool correlating function to brain plasticity.

\end{abstract}

\section*{KEYWORDS}
Unsupervised behaviour analysis; deep learning; computer vision; artificial intelligence; sensorymotor behaviour and recovery; automatic quantification and diagnostics of sensorymotor deficits

\section*{INTRODUCTION}
Motor behaviour, i.e. the dynamic change of posture, is the visible result of elaborate internal processes. The precise analysis of motor behaviour and its deviations, consequently, constitutes an essential, non-invasive diagnostic strategy\cite{measuringbehavior2018} in many fields ranging from biomedical fundamental research on animals to clinical diagnosis of patients. Behaviour is the output of coordinated internal processes of the brain including the planning and fine tuning of movements in higher brain regions, trans-ducting the signal to the spinal cord and converting it to an orchestrated muscle activation for complex movements.
A detailed analysis of skilled behaviour and its impairment is, therefore, crucial for the neuroscientific understanding of brain (dys-)function. Moreover, the objective quantification of motor impairment in patients is not only valuable to detect and classify distinct functional deficits. It can also serve as basis for individually optimised treatment strategies\cite{filli2018profiling}. Videos of behaviour recorded during the long-term recovery after neurological diseases provide an easily available, rich source of information to evaluate and adjust drug application and rehabilitative treatment paradigms.\\

The main bottleneck in behavioural studies is presently that all analysis of skilled motor function depends on a time-intensive, potentially subjective, and costly manual evaluation of behaviour: Behaviour analysis has so far mainly relied on reflective physical markers placed on body joints \cite{measuringbehavior2018,graspfrommotorcortex2010,moshBlack,mocapBlack} or on tracking manually labelled virtual keypoints in video recordings of behaviour \cite{machine_vision_methods, automated_imagebased_tracking, ratsocialtracking2016, deepbehavior, deepposekit, leap}.
However, placing physical markers can be tedious and distracting, especially when working with animals. In contrast, virtual keypoints are beneficial due to their non-invasiveness, but they require substantial effort for keypoint annotation. To avoid labelling every video frame, machine learning has been employed to automatically track body-parts\cite{deeplabcut, deepbehavior, deepposekit, leap, simon2017hand}. 
For example, DeepLabCut\cite{deeplabcut} has been successfully applied to different experimental settings. Mathis et al.\cite{nath2019using, mathis2020deep} show how this algorithm can be utilised for different species and behaviours. However, applying a keypoint model to novel datasets requires fine-tuning based on extra manual annotation for the specific data.
Where such manual annotation is not an issue or for data for which existing models already work sufficiently well, keypoint approaches offer a simple and effective solution. 
This is also reflected by the recent work in the computer vision community on keypoint based models for animal posture estimation \cite{mu2020learning, li2020deformation} or on keypoint-based inference of three dimensional human body models that is not generic in its applicability to other species \cite{sanakoyeu2020transferring, kocabas2020vibe, SMPL2015, Zuffi19Safari, habermann2020deepcap}.
However, the great benefit of simplicity of a posture representation based on keypoints limits a detailed analysis of arbitrary novel, e.g. impaired, behaviour: A detailed representation of a priori unknown body movements requires trained keypoints for almost every body joint, which presents an impracticable effort to supervised training. Therefore, users have to limit the keypoints to a predefined feasible subset. We argue that this introduces a problem: to select the optimal keypoints for a detailed analysis, the user needs to know what the behaviour of interest is \emph{before} applying the posture detection.
However, a true diagnostic tool should \emph{discover} and localise deviant behaviour, rather than \emph{only confirm} it. Consequently, there is a human annotator bias: the behaviour analysis is restricted to the keypoints that a user has decided for and different annotators may favour different points. Thus, it is missing features that may be relevant to fully characterise motor behaviour and draw appropriate conclusions on underlying neuronal mechanisms.
Several recent works on behaviour analysis also confirm these drawbacks of using a parametric model, such as the loss of information \cite{nips_behavenet}, and, thus, propose approaches using non-parametric models that avoid the aforementioned prior assumptions on the data \cite{nips_behavenet, ryait2019plosbio}.
Compared to their method, our model is also able to compare behaviour \emph{across different} subjects and over time, moreover we can identify and visually magnify the movement deviation between subjects.

We propose a fully automatic, unsupervised diagnostic support system for behaviour analysis that
can discover even subtle deviations of motor function. The approach not only extracts and classifies behaviour \cite{jaaba, ryait2019plosbio}, but it can also compare and quantify even small differences. Neither analysis of novel video sequences nor training of the underlying deep neural network require physical markers or supervision with tedious keypoint annotations.
This avoids a user bias of having to select appropriate keypoints for training a keypoint model and also supports an objective analysis.
Our approach automatically discovers characteristic behaviour, localises it temporally and spatially, and, above all, provides a behaviour magnification that not just highlights but also amplifies subtle differences in behaviour directly in the video: Given a novel video sequence, the approach can automatically compare the behaviour against reference videos showing healthy or impaired behaviour of other individuals, since it is invariant to inter-subject variations in appearance. Also, behaviour can be contrasted against previous videos of the same individual during a recovery process to identify the fine differences. Behaviour magnification then uses a generative neural network to synthesise a new video with the subtle deviations between healthy and impaired being amplified so they become clearly visible.
Key to our model is a disentangling of posture and appearance for image synthesis to amplify only the deviations in behaviour across individuals despite differences in appearance.
We assume a clinical or research setting with static background and controlled recording. Disentangling should not merely separate moving from static image regions. Otherwise we would merge non-moving body parts with the background, hindering analysis and magnification across different subjects. Rather we need to learn the commonalities of reference behaviour across different subjects and disentangle this from their individual deviations in appearance.
Our algorithm is promising for diverse applications in the field of biomedical research and was evaluated on rodents and human patients with disease models such as stroke and multiple sclerosis.\\

Our \textit{uBAM} (unsupervised behaviour analysis and magnification) interface is freely available online and interactive demos are provided on \\ \url{https://utabuechler.github.io/behaviourAnalysis/}.

\section*{UNSUPERVISED METHOD FOR BEHAVIOUR ANALYSIS AND MAGNIFICATION}
Here, we first discuss our two main methodological contributions with their possible applications: a neural network for comparative analysis of behaviour and one for magnification of differences in behaviour.

\subsection{A Convolutional Neural Network for Comparing behaviour}
To train a model for an objective behaviour analysis, an unsupervised learning procedure (Fig. 1) is crucial to circumvent tedious annotations and an annotator bias. The challenge is to learn a representation that can compare similar behaviour across individuals despite their difference in appearance (discussed subsequently in Fig. 2a,b). 
We present a deep neural network that extracts the characteristic behaviour from a video sequence without any user intervention (Fig. 1a). To train this model, we propose a surrogate task, based on \cite{lstm2017cvpr,irl2018cvpr,jigsaw,opn}, where the network learns to distinguish a behaviour sequence $x=(x_1, x_2,...)$ from the same sequence with frames $x_{\rho_{(i)}}$ randomly permuted by a permutation $\rho$ (Fig 1b). Solving this auxiliary task requires the network to learn a representation of posture $F_\pi(x_i)$ and its temporal dynamics, the behaviour $F_\beta(x)$.
We test our posture and behaviour representation qualitatively in Fig. 2a,b and quantitatively in Supplementary Tables 1-4.
Moreover, our prior work\cite{lstm2017cvpr} provides an evaluation of the proposed learning strategy from a computer vision perspective by also testing on several standard human posture benchmarks.

\begin{figure}
\centering
\includegraphics[width=0.95\linewidth]{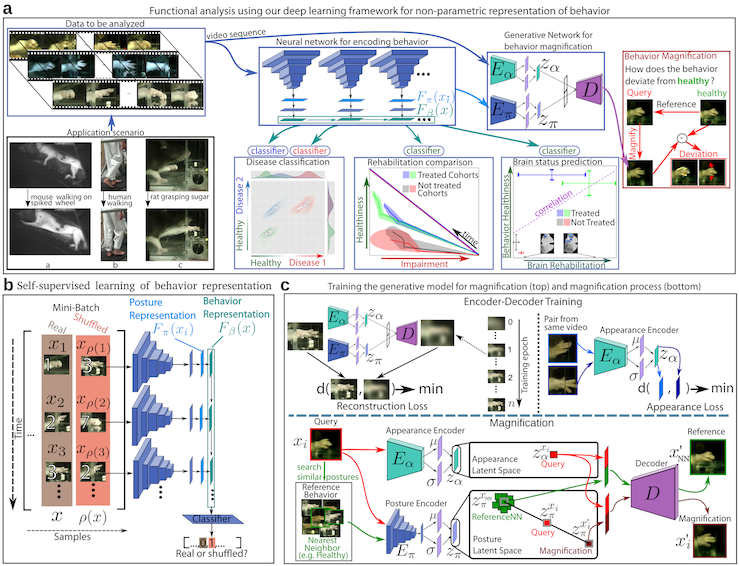}
\caption{
\textbf{Unsupervised behaviour Analysis and Magnification. } 
\textbf{a,} Our framework for analysing behaviour. (center top) the proposed neural network for behaviour encoding and magnification. (center-bottom and right) examples of the applications enabled by our approach. (left) Exemplary frames of the three different species we evaluate our method upon: mouse walking on a spiked wheel, humans walking on a treadmill, rats grasping sugar. 
\textbf{b,} The proposed procedure for learning the \textit{behaviour representation} without manual annotation. The network is trained to extract a posture and behaviour embedding ($F_\pi(\bullet)$ and $F_\beta(\bullet)$) by forcing it to distinguish the characteristic structure of behaviour sequences from shuffled sequences.
\textbf{c,} Training our encoder-decoder model to disentangle posture and appearance for behaviour magnification. (top-left) Our reconstruction loss to train the generative model. (top-right) The appearance encoder is trained using our \textit{appearance loss} that maps different postures from a video to the same embedding vector. (bottom) behaviour magnification: we can magnify the difference between a query posture to a reference behaviour by extrapolating the corresponding point in the posture latent space $z_\pi$. The reference behaviour is described by a collection of postures from a reference of interest (e.g. healthy). The magnified image is generated by combining the query appearance with the magnified posture.
}
\end{figure}

\begin{figure}
\centering
\includegraphics[width=0.95\linewidth]{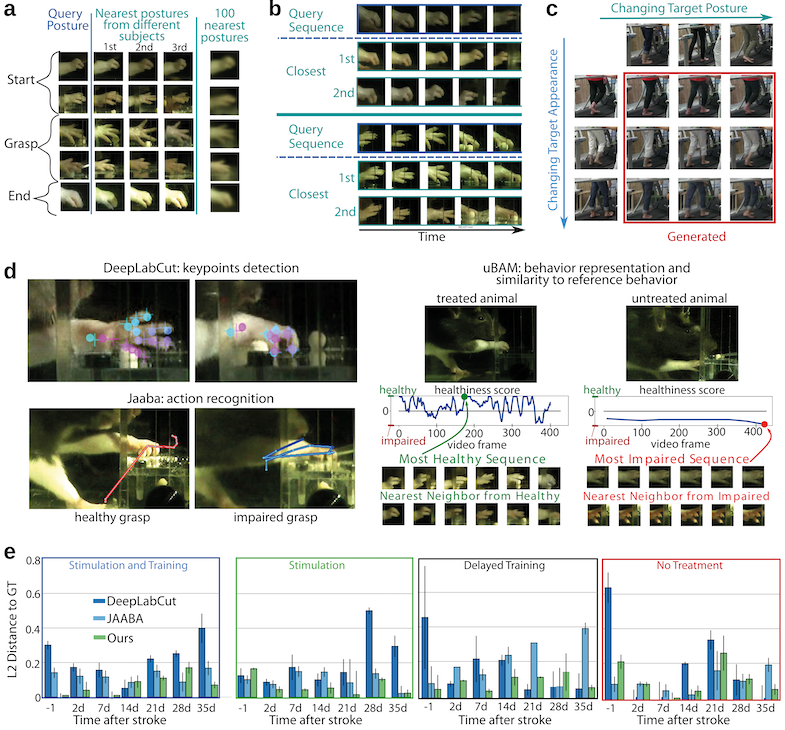}
\caption{
\textbf{Evaluating the learned disentangled representation.}
\textbf{a,} Three nearest neighbors for different query frames based on our unsupervised \textit{posture} representation. Averaging the 100 nearest neighbors (right column) demonstrates congruency in posture.
\textbf{b,} Two nearest neighbors for query sequences based on our unsupervised \textit{behavior} representation.
\textbf{a-b,} The representation successfully disregards appearance differences and learns to measure similarity solely based on posture and behavior.
\textbf{c,} Our encoder-decoder model successfully transfers the posture of a subject to another while keeping the appearance unaltered, showing a good separation of posture and appearance encodings.
\textbf{d,} Current methods for behavior analysis. (top-left) DeepLabCut$^{11}$ is a deep neural network for key-point detection. (bottom-left) JAABA$^{13}$ is a tool for behavior classification based on key-point kinematics. (right) Our approach learns a representation of behavior which enables comparison of each sequence by identifying most related reference behavior (healthy or impaired).
\textbf{e,} Comparing our non-parametric model and classical trajectory based methods.
The bar-plot shows the euclidean distance from the healthiness ground-truth (GT)$^{42}$ score provided by the experts (\textit{lower is better}). For each model, the experiment is repeated on different train/test split, summarised by the standard deviation. The results of our model correlate with GT at $0.933\pm 0.005$, while DeepLabCut and Jaaba are at $0.820\pm0.072$ and $0.695\pm0.014$, respectively. 
}
\end{figure}

\subsection{A Generative Neural Network for Behaviour Magnification}
True diagnostics requires not only to discover and quantify impairment. The user also needs to understand \emph{how} a behaviour deviates from some reference behaviour (e.g. healthy) to interpret the symptoms and actually understand the disorder. Thus, small deviations in behaviour that might easily be overlooked need to be magnified.

Existing methods on video magnification have mainly addressed the magnification of small motions \cite{deepmag, magnification, wu2012eulerian, elgharib2015video, wadhwa2013phase, wadhwa2014riesz, zhang2017video, tulyakov2016self, revealingnonlocal} or the deviation from a predefined reference shape \cite{deviationmagn}, but only within the same video \cite{magnification,deepmag}.
We propose an approach for magnifying differences in behaviour even across individuals (Fig. 1c). To ease interpretation for the user, we amplify characteristic differences directly in the query video by altering its individual frames to magnify their deviation in posture from the reference. This requires to re-synthesise one subject with the appearance of another to enable a direct visual comparison of their posture (Fig. 2c) and to visually amplify the differences (discussed subsequently in section Behaviour magnification).
Generative models have been very successful in image synthesis. However, standard generative models (VAE\cite{vae}, GAN\cite{gan}) in their basic form lack the ability to disentangle aspects that need to be magnified (characteristic deviations in posture) from those that should not be altered (appearance). 

Therefore, we need a model that learns directly on videos of reference subjects to disentangle\cite{vunet} posture $E_{\pi}(x_i)$ from appearance $E_{\alpha}(x_i)$ (for details see Methods).
A suitable generative model should be able to combine any posture and appearance pairs and reconstruct a new image even if this specific combination does not exist in the dataset. In Extended Data Figure 1, 2 and 3, we compare our method with previous magnification methods.

Our unsupervised posture representation (previous section) is key for the disentanglement since it extracts a posture representation $F_\pi$ not biased by the appearance and is used to produce $E_\pi(x_i) = f(F_\pi(x_i))$ where $f(\bullet)$ is a linear layer. The trained model can then be applied to amplify behaviour deviation of a novel query sequence against reference videos (see Method).

\section*{EXPERIMENTAL EVALUATION}
In this section, we evaluate our uBAM framework on different species and tasks, from subject rehabilitation to brain status correlation.

\subsection{Experimental Setup for Benchmarking.}
We evaluated our framework on three different species studying diverse motor functions (Fig. 1a left). Videos were recorded using a single consumer camcorder to demonstrate the low-cost and simple applicability of our approach to behaviour analysis. In Supplementary we describe the acquisition process.

Firstly, we analysed the recovery of impaired forelimb function in a rat stroke model where a stroke partially destroyed the sensorimotor cortex in one hemisphere. The resulting impact on skilled motor function was assessed in several animals grasping for a sugar pellet through a small opening. The animals were randomised in four different treatment groups (see Supplementary). Recordings (50 frames/sec, consumer camera) have been taken during the initial training of the animals before stroke and during recovery, which lasted up to 5 weeks after stroke. 

Secondly, we trained mice to run on a wheel with irregular rung distance while being head-fixed. Animals were recorded (100 frames/sec with an infrared camera) before and after triggering a stroke which compromises their ability to precisely target the rungs of the wheel.

Thirdly, healthy human subjects and patients with different neurological disorders were recorded while walking on a treadmill. Patients with multiple sclerosis (MS) and hydrocephalus (HC) were analysed. MS subjects receiving a dose of fampyra and hydrocephalus patients with lumbar puncture were filmed with a standard consumer video-camera on the treadmill again 24h after the treatment.
Both are routine interventions in the clinical setting and have been associated with improvements of ambulatory function\cite{goodman2009sustained,zorner2016prolonged,schniepp2017walking}.

\subsection{Evaluation of Behaviour Encoding. }
We first evaluated the capability of our learned representation (Fig. 1b, Extended Data Figure 1 and Supplementary Figure 1) to compare postures and behaviour sequences across different individuals. Therefore, we use the posture encoder to measure the similarity between a query frame and frames from videos showing other rats grasping for a sugar pellet. Fig. 2a presents the three nearest neighbours and an average over the hundred closest samples. Evidently, despite variations in appearance, similar postures are identified (see Supplementary Tab.1 for additional evaluation on the posture representation). In Fig. 2b for a query sequence, sub-sequences with similar behaviour are extracted from the videos of other rats. In Extended Data Figure 4 we compare our posture representation against a classical projection method.

\subsection{Invariance to Appearance Variation.} Next, we investigated the invariance of our posture representation against appearance changes. Based on both disentangled encodings, $E_\alpha$ and $E_\pi$, we utilise our generative network from Fig. 1c to extract the posture from a frame and synthesise it with the appearance of another individual. Fig. 2c shows that posture is correctly preserved along columns and appearance along rows. This clearly demonstrates that the approach can analyse posture across individuals despite strong variations in appearance, which is also key to behaviour magnification (see also Extended Data Figure 6).

\subsection{Comparing Against Keypoint-based Methods.} 
We compared two currently popular approaches for supervised, keypoint-based behaviour analysis (Fig. 2d), DeepLabCut\cite{deeplabcut} and JAABA \cite{jaaba}, against our approach which requires no posture annotation.

Fig. 2e shows the results of all three approaches on video data of rats grasping a sugar pellet. The data includes four cohorts of rats undergoing different rehabilitative treatments during a 35-day long recovery.
Each bar in Fig. 2e represents the error from ground truth provided by experts (see Methods). 
In Extended Data Figure 8, we compare our representation with a state-of-the-art deep neural network specific for action recognition\cite{r3d}.

The results produced by our model have a correlation of $0.933\pm 0.005$ with the ground truth scores annotated by neuroscientists. Results from JAABA and DeepLabCut are $0.820\pm0.072$ and $0.695\pm0.014$, respectively. The analysis shows that, compared to keypoint-based methods, our non-parametric model can extract more detailed information necessary for a fine-grained study of motor function, while avoiding tedious manual annotations. 

\subsection{Behaviour-based Classification of Diseases}

Next, we evaluated the capability of our approach to distinguish different diseases solely based on behaviour analysis. As in the previous paragraph we also trained a linear discriminative classifier to distinguish healthy from impaired sequences. On the test data we obtained an accuracy of $88\pm11\%$ for the mice (intact vs. stroke) and $90\pm9\%$ for the humans (healthy vs. hydrocephalus/MS). 

Furthermore, Fig. 3a shows our unsupervised behaviour representation $F_\beta$ from Fig. 1b directly mapped onto two dimensions using t-SNE \cite{tsne} for the dataset of mice running on the wheel. Without any annotation of impairment during training, the behaviour encoding still captures important characteristics to separate healthy from impaired.

\begin{figure}
\centering
\includegraphics[width=0.85\linewidth]{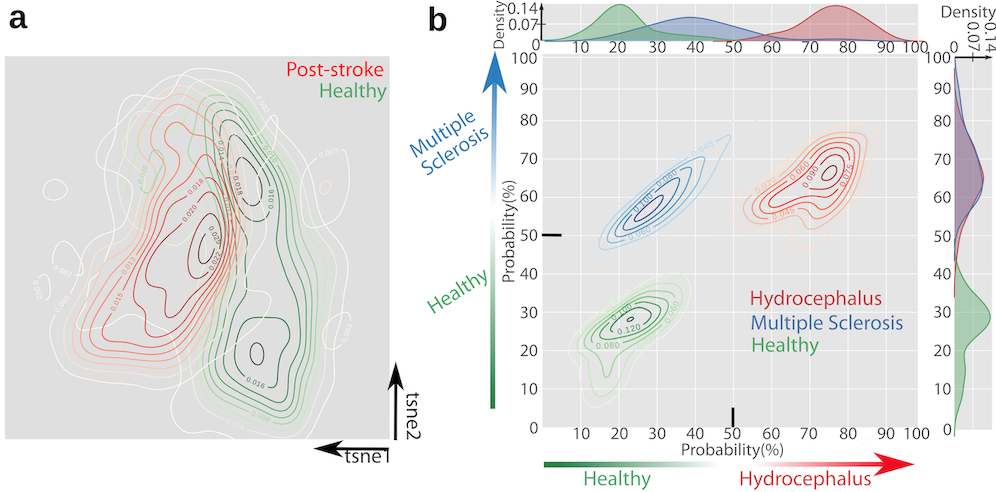}
\caption{
\textbf{Behavior analysis for disease classification.}
\textbf{a,} 2D visualization of our behavior representation for mice using tSNE$^{37}$.
Although no supervision has been used, the two distributions, pre-stroke (healthy) and post-stroke (impaired), are well separated.
\textbf{b,} Predicting the disease of human patients using our behavior representation. Each axis gives the probability of a subject to be affected by the respective disease or not. The plot shows that the three disease cohorts are well separated and compact, indicating a decent separation based on behavior analysis.
}
\end{figure}

Moreover, we studied whether our approach can distinguish different diseases from another (Fig. 3b). As beforehand, for the human gait data we employ classifiers on top of our behaviour representation. Based on gait we train one classifier to recognise patients with MS and another for patients with hydrocephalus. The successful separation of the different behaviours by a simple linear classifier is only possible due to the detailed behaviour encoding extracted by our approach.

\subsection{Visualising the acquirement of skilled motor function}

Our approach can also quantify and visualise the subtle changes in behaviour while animals learn to perform skilled motor function, such as rats learning to grasp a sugar pellet through a slit. Fig. 4a (top) compares the grasping during 31 days of learning against the skilled behaviour acquired at the end of training, which serves as reference. Additionally, we can identify for each time point of learning how the behaviour differs from the reference: the posture representation allows to spot postures (rows in Fig 4a bottom) that are substantially over-represented in contrast to the skilled reference (red) as well as the ones that are missing (blue) at each point in time. Here the postures are mapped from the multidimensional $F_\pi$ to 1D on the y-axis using t-SNE. 
The result shows that non-grasping postures (bottom) are more frequent in early stages, while grasping postures (top), which precisely target the sugar pellet, are unlikely. 
During learning, the posture distribution then converges to the skilled reference.

\begin{figure}
\centering
\includegraphics[width=0.99\linewidth]{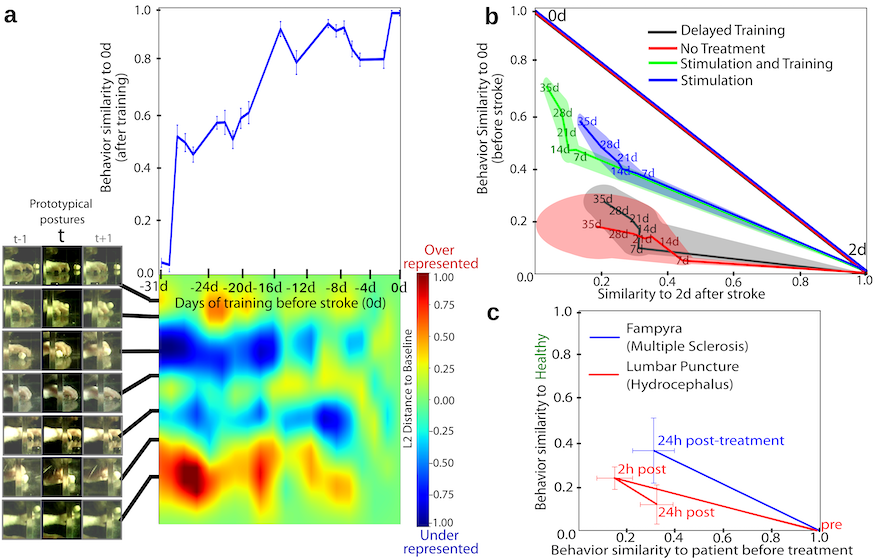}
\caption{
\textbf{Evaluation of motor function skills at different points during learning.}
\textbf{a,} Posture analysis of rats during initial training of grasping.
(top) Similarity to trained animals (0d). The similarity gradually increases over successive days.
(bottom) Relative frequency of individual postures per day of training compared to trained baseline (BL) animals by Euclidean distance (L2). green: same frequency, red: more frequent, blue: less frequent. 
Examples of postures are shown on the left ($t$ column), including temporal context ($t-1$ and $t+1$ columns) to better understand the ordering of the postures along the vertical axis. It shows that correct hand closure at grasping occurs notably less early in training.
\textbf{b,} Comparison of different treatments on the rats dataset during 35 days of rehabilitation. The axes measure similarity to healthy (pre-stroke, vertical, same axes as curve in a-top) and impaired (2days post-stroke, horizontal). Rehabilitation successfully restores motor function to bring behaviour close to pre-stroke for the treatment cohorts (green, blue). Without treatment (black, red cohort), behaviour is altered, but still remains far from pre-stroke, indicating inadequate compensation of motor-function.
\textbf{c,} Behaviour analysis of human patients after different disease specific treatments.
For both groups the similarity to healthy subjects increases after the treatment, indicating an improvement in behaviour. Our evaluation confirms the supposition of experts that the motor skills of LP treated patients deteriorate again within $24$ hours. For all plots, the standard deviation is computed by repeating the experiment on different train/test splits.
}
\end{figure}

\subsection{Behaviour-based Evaluation of Different Therapies}

Next, we compared different therapies after stroke by measuring the change in behaviour they cause in the grasping task. Fig 4b shows the similarity per rat cohort to a healthy baseline (y-axis) and to immediately post-stroke (x-axis) for each week of recovery. The cohorts with optogenetic stimulation of the intact pyramidal tract ("brain stimulation" groups in green, blue) steadily improve during rehabilitation, reaching around $70\pm3\%$ similarity to the baseline behaviour and having almost no similarity (green $5\pm3\%$, blue $15\pm5\%$) with the post-stroke behaviour. 
In contrast, groups with no treatment or only rehabilitative training in the grasping task reveal behaviour that is similar to neither reference ($<35\%$), suggesting an inadequate compensation differing substantially from true recovery of impaired function.

Fig 4c depicts gait improvement for patients after a disease specific treatment as similarity to pre-treatment (x-axis) and to behaviour of healthy subjects (y-axis). Fampyra, a reversible potassium channel blocker applied over a period of 14 days (2x10mg per day) yields a substantial improvement on the walking function of MS patients. Lumbar puncture leads to a major improvement within the first 2 hours for patients affected by hydrocephalus - an effect that only initially lasted for 2h due to the reduction in cerebrospinal fluid pressure.

The experiments show the behaviour encoding $F_\beta$ to be an effective tool for both, to compare different therapies after a disease and to diagnose the resulting changes in motor function.

\subsection{Analysing Relations Between Behaviour and Neuronal Rewiring}
Our behaviour representation is sufficiently sensitive to discover a correlation between restoration of behaviour and cortical rewiring in the impaired motor cortex.

For each treatment cohort (from Fig. 4b), Fig. 5a compares the degree of recovery by comparing the behaviour 35d post stroke against the healthy pre-stroke baseline.
In addition, we counted out-sprouting nerve fibers (BDA positive, see Methods) in the peri-infarct cortex post mortem as a characteristic of plastic rewiring processes in the impaired hemisphere which are influenced by the post-stroke treatment. The plot shows a significant correlation of $r \sim0.7$ between our measured behaviour changes and the degree of neuronal fiber sprouting in the perilesional cortex. Hence, our non-invasive behaviour analysis is discriminative enough to provide a measurement of the degree of recovery for impaired motor function that relates to the underlying subtle neuroanatomical modifications in the brain.

\begin{figure}[b]
\centering
\includegraphics[width=0.75\linewidth]{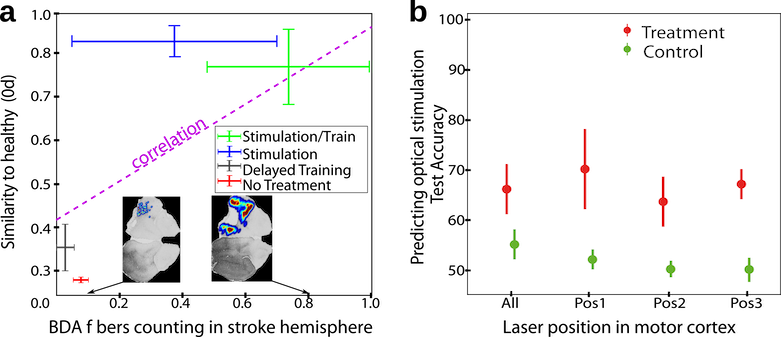}
\caption{
\textbf{Predicting neurophysiological characteristics from behavior.} 
\textbf{a,} Relation between cortical rewiring (number of Biotin Dextran Amine (BDA)-positive fibers post-mortem in the hemisphere affected by the stroke, horizontal axis) and our behaviour representation (vertical). Two examples, low and high fiber density, are shown.
The experiment indicates that the behaviour and rewiring are correlated ($p\sim0.7$).
\textbf{b,} In-vivo deactivation of cortical circuits using optogenetic stimulation triggers subtle, reversible changes in behaviour. Only based on our behaviour representation, a classifier is then able to predict for test sequences whether there was optical stimulation. As would be expected, the classifier only achieves chance level for a control cohort without neuronal silencer but it performs substantially better ($68\pm9$\% accuracy) on the treated animals. (Details on the laser positions are given in the online Methods). \textbf{a, b,} The standard deviation is computed by averaging across different animals.
}
\end{figure}

\subsection{Simultaneous Optogenetic Stimulation and Behaviour Analysis}

Optogenetics allows to reversibly deactivate cortical circuits in-vivo.
We expressed the optical neural inhibitor ArchT in a subset of corticospinal projecting neurons in rats and compared their grasping skills with and without light-induced silencing of these distinct neurons. Control animals were treated in the same way, however lacked the expression of the neuronal silencer ArchT. 
Extended Data Figure 10 shows a grasp example for light on and light off selected using our classifier.
As expected, the classifier (Fig. 5b) only performs at chance level for controls and is substantially better ($68\pm9\%$ test accuracy per grasp) discriminating light-perturbed behaviour in animals expressing ArchT in parts of the pyramidal tract. Only for these it can recognise the light-driven modification of specific motor functions. Since such altered behaviour is not present in every try of a grasp, the goal had to be a substantial improvement over control, but not finding differences in every grasp.
In fact, depending on the potency of the inhibitory opsin used and the specific neurons being silenced, optogenetic inhibition may not alter behaviour in every trial due to many compensating mechanisms. This effect for the opsin ArchT which we used here was also found by other groups in other contexts \cite{lafferty2020off, miao2015hippocampal, carta2019cerebellar}.

\subsection{Behaviour Magnification}\label{sec:magnification}

Finally, we tested our behaviour magnification as a diagnostic tool to discover variations in posture between a query and reference posture. 
The results demonstrate that differences barely visible between query and reference are correctly magnified by our model and can be spotted and understood more easily with the help of our highlighting and magnification. 
Notice that our approach does not magnify arbitrary deviations caused by differences in appearance or normal posture variations, but only those due to impaired behaviour as analysed in Extended Data Figure 2 and 3. 

Consequently, our behaviour magnification can extend medical diagnostics to reveal deviations due to impairment that even the trained eye might overlook.
For example, the first subject in Fig. 6b seems to show no impairment during the step, however the magnified posture reveals that the left steps are too short compared to the healthy references. The second subject has no problems with the right step, but shows difficulties in keeping the left leg straight at the beginning of the left step. The magnification in Fig. 6c reveals that the rats have especially problems with the correct supination of the paw when approaching the sugar.

\begin{figure}
\centering
\includegraphics[width=0.90\linewidth]{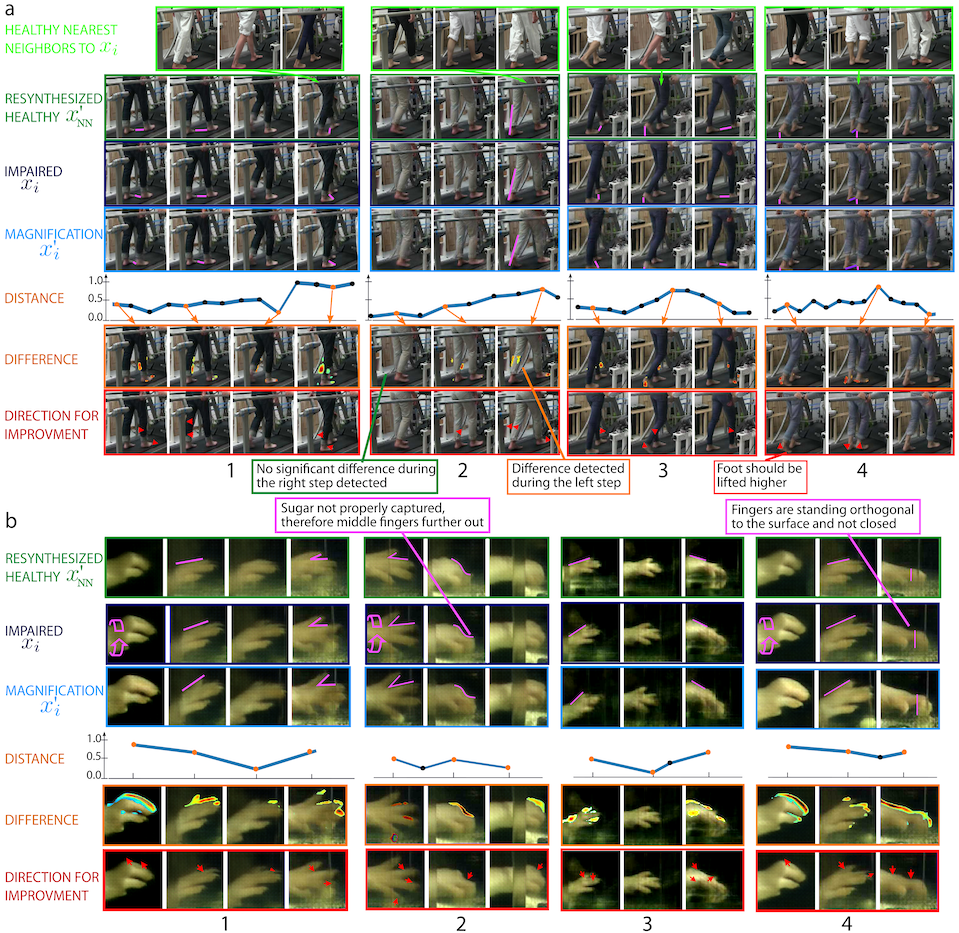}
\caption{
\textbf{Magnifying impaired behavior as a diagnostic tool.}
Our approach magnifies and localizes the differences between a query sequence $x_i$ and a reference of closest postures from healthy subjects (top row) that are averaged. To support a direct diagnostic comparison despite their difference in appearance, we re-synthesize the healthy $x_{NN}^\prime$ with the query appearance $z_\alpha^{x_i}$. Our magnification $x_{i}^{\prime}$ reveals the hardly visible differences in posture between the query $x_i$ and reference $x^{\prime}_{NN}$. Neurologists superimposed magenta markers in the first three rows showing their subsequent diagnostic process. Our framework highlights the \emph{differences} between healthy and impaired by a heat map.
\emph{Distance} shows overall posture deviation per frame of magnification from healthy to find impairment. \emph{Direction for Improvement} shows how the respective body region needs to move to compensate the impairment.
\textbf{a,} Examples of an analysis on the dataset of human patients. \textbf{a.1} The patient performs smaller steps with right leg; the \emph{Direction for Improvement} indicates that extending the right leg during the right step would yield an improvement. \textbf{a.2} The patient shows no issues during the right step, while the left leg cannot be bent properly. \textbf{a.3-4} An improvement can be achieved by further lifting up the right heel.
\textbf{b,} Exemplary results on the rat stroke dataset. In \textbf{b.1,3 and 4} the paws are more pitched upwards compared to the healthy when approaching the sugar. Moreover, when initiating the grasp in \textbf{b.1,2 and 4}, the impaired paws show too much supination.
}
\end{figure}

\section*{DISCUSSION}

A detailed analysis of motor behaviour and its impairment is crucial as a basis for understanding neural function. Current methods represent behaviour as trajectories of keypoints, thus reducing complex motor kinematics to few points that must be provided beforehand. Rather than supporting users with an automatic discovery of deviant behaviour, users need to know its characteristics before performing the analysis to define it in terms of keypoints.

We present a self-supervised deep learning approach for training a convolutional neural network that provides a fully automatic, unsupervised diagnostic behaviour analysis. A second generative network helps to interpret this analysis by magnifying characteristic differences in behaviour even across individuals. The method is provided as an open source software package called \textit{uBAM}.

A possibly limiting factor of our method is the requirement of a controlled recording environment: the camera view should be roughly the same across subjects and the background mostly static. We believe these constraints to be reasonable for biomedical research and common in the lab and clinical practice. Nevertheless, our videos feature some changes in the camera viewpoint and contain other moving objects (e.g. nurses in the human data), thereby proving our model to be robust to small changes.

In order to retrieve an evaluation focused solely on the impaired subject, our model utilises a pre-processing step for extracting the corresponding region-of-interest. For our experiments we utilised simple techniques that proved to be robust in the standard clinical and lab setting. However, since our approach is independent from this pre-processing, the user is free to utilise other, potentially even supervised detection methods for their data.

This paper focused solely on analysing the behaviour of individual subjects. A further extension to study the interaction of multiple subjects would have wide applicability\cite{jaaba}, especially when investigating animal social behaviour. Moreover, it would be interesting to analyse the relation of behaviour to complex brain activity in vivo across diverse brain areas, rather than only in selected subareas of the motor cortex.

Our method is not meant to completely eliminate key-point based methods, but it rather is of complementary nature. Besides being feasible where physical markers would not or where training virtual markers is too costly, our approach can discover deviant behaviour, which is challenging with user-defined key-points that would require the relevant body-parts to be known and chosen before the actual diagnosis. After using our model for discovering the affected body-part(s), the user can complement the analysis by incorporating further studies based on virtual markers which are optimised for the affected body-parts.

Combined behaviour analysis and magnification together provide a powerful, non-invasive approach to not only support a biomedical diagnosis, but also to aid its interpretation. Moreover, due to the learned invariance to appearance differences, behaviour can be objectively compared across individuals. We believe that the proposed efficient, unsupervised, keypoint-free behaviour analysis provides a generic foundation to explore diverse novel applications in the field of biomedical research, non-invasive diagnostics, and beyond. The proposed system can assist in vivo biological experimentation and has the potential to support a better understanding of neuronal processes in the brain, their degeneration, and to discover new treatments.

%
%
%
%
%
%
%
%
\clearpage
\section*{METHOD DETAILS}

\subsection{Encoding the Behaviour Representation.}

We propose a non-parametric model to extract the essential information for representing behaviour necessary for several biomedical analysis tasks (Fig. 1a middle). This deep neural network based encoding is trained using our unsupervised process summarised in Fig. 1b. Through a surrogate task we circumvent the need for manually annotated labels. For each sequence $x = (x_{1}, x_{2}, ...)$ of frames $x_{i}$, a sequence $\hat{x} = (x_{\rho(1)},x_{\rho(2)}, ...)$ is constructed by random permutation $\rho$ of the frames. The network is trained to distinguish between real and permuted sequence. Solving this surrogate task, i.e. identifying the real sequence, allows the network to learn the correct execution of a movement, which is defined by the change of posture over time (behavior). 
Moreover, to solve the surrogate task, the network needs to ignore the appearance within frames because every frame in a sequence $x$ has the same appearance as the frames in a shuffled sequence $\hat{x}$, thus appearance is not a discriminative characteristic between the two sequences. Keeping information about appearance can only hurt the model, so the network learns to be invariant to that. On the other hand, given that nearby frames $x_{i}$ and $x_{i+1}$ can have only slightly different postures, the network needs to learn a fine-grained posture representation to predict the order of two consecutive frames.
The network trained using our surrogate task produces then a posture encoding $F_\pi(x_i)$ and behaviour encoding $F_\beta(x)$ used in our experiments.
Our model is trained on unlabelled videos from both healthy and impaired subjects. Therefore, as shown in Fig. 2a,b, the learned representation is invariant to inter-subject differences in appearance and can be employed for different types of impairment. On top of this encoding we can now train simple classifiers $f(F_\beta(x))$ to recognise impaired behaviour, to distinguish different diseases from another (Fig. 3), and to measure the degree of recovery during rehabilitation (Fig. 4). These downstream classifiers are superimposed on top of our unsupervised behaviour representation and only utilise video meta information (such as time after stroke), but no manual annotation.

\textit{Architecture for Behaviour Representation.} Our unsupervised training algorithm is independent from the underling network architecture, allowing to easily change architectures. We use AlexNet\cite{alexnet} because it performs relatively well on diverse tasks and is fast to train with a single GPU. Further information can be found in the Supplementary material.

\textit{Unsupervised Training of the Generative Model.}
Behaviour magnification demands an approach for image generation that learns to alter posture independently from appearance. 
The most popular generative models are currently Generative Adversarial Networks (GANs), since they can produce very realistic images. 
However, GANs are not suited for magnification because they synthesise new images starting from a random vector instead of an input image.
Therefore, we base our model on the autoencoder\cite{autoencoder} (AE) which reconstructs the image given as input to the network. An AE encodes the image into a low dimensional embedding which is then decoded back to the pixel space.
However, a classical AE does not explicitly disentangle appearance and posture. Hence, we propose our AE composed of two encoders $E_\alpha$ and $E_\pi$ to describe a frame $x_i$ by the appearance $E_\alpha(x_i)$ and posture $E_\pi(x_i)$ of the subject shown in $x_i$. $E_\alpha$ and $E_\pi$ encode the image $x_i$ into the low dimensional space $z_\alpha$ and $z_\pi$, respectively. Given the two encodings, the decoder $D$ is used to reconstruct the input frame $x_i$ and outputs the reconstruction $x^\prime_i = D(E_\pi(x_i),E_\alpha(x_i))$ (Fig. 1c top-left). To produce realistically-looking images, our model is trained by minimising the reconstruction loss $\mathcal{L}_{rec} = d(x_i,x^\prime_i)$, which computes the distance between synthesis $x^\prime_i$ and the input image $x_i$ with $d(\cdot,\cdot)$ being a standard perceptual loss\cite{perceptualLoss}, i.e. based on VGG pre-trained on ImageNet. We found empirically that the perceptual loss produces sharper images in contrast to the pixel-wise L2 loss. As a side note, it has been shown that, in some cases, perceptual loss might introduce artefacts in the generated images given by the different distributions between source (ImageNet) and target data \cite{lafferty2020off}. However, we did not experience such artefacts on our generated images.

Additionally, we introduce a variational factor\cite{vae} into our generative model, which has been proven to be beneficial for improving the visual quality\cite{vae}. During training, the encoder outputs the mean and standard deviation (std) for a normal distribution $z \sim \mathcal{N}(\mu,\,\sigma^{2})$, from which the encoding vector $z$ is sampled for synthesising the image. To guarantee the sampling distribution to be normal, the Kullback–Leibler distance $\mathcal{L}_{KL}$ between encoder output and normal distribution is minimised.

We enforce the \textit{disentanglement} of posture and appearance by introducing two additional training objectives that determine each encoder. 
We leverage the posture encoder $F_\pi$ from the surrogate task to produce the encoder $E_\pi$ of the generative model. $F_\pi$ is learned beforehand using our unsupervised behaviour representation (Fig. 1b) which assures the encoding to only retain posture information. Nevertheless, a linear mapping is needed to incorporate $F_\pi$ into the generative model for two reasons: reducing the dimensionality to create the bottleneck effect typical of auto-encoders; and to learn to describe a gaussian distribution through the KL-loss. Thus, we define $E_\pi(x_i) = f(F_\pi(x_i))$ by applying a linear layer $f(\bullet)$ on top of our posture representation $F_\pi(x_i)$ to produce a posture encoder for disentangling appearance from posture. While the representation $F_\pi(x_i)$ is fixed during the VAE training, the linear layer $f(\bullet)$ is trained with the reconstruction loss $\mathcal{L}_{rec}$ and $\mathcal{L}_{KL}$.
For the appearance encoder $E_\alpha$ we propose the appearance loss $L_{app} = | E_\alpha(x_i) - E_\alpha(x_j) |^2$ with $x_i$ and $x_j$ two frames from the same video thus having the same appearance. This loss ensures a similar appearance encoding despite differences in posture between the two frames (Fig. 1c top-right).

Combining the above losses yields the following training objective for our generative model,
\begin{align}
    \mathcal{L} &= \mathcal{L}_{rec} + \gamma \mathcal{L}_{app} + \eta \mathcal{L}_{KL},
\end{align}
where $\gamma$ and $\eta$ are free parameters, with $\gamma = \eta = 10^{-3}$ in our experiments. The details of the autoencoder architecture can be found in the Supplementary material.

\textit{Evaluating the Disentanglement of Posture and Appearance.} After training, we evaluated the ability of the generative model to combine appearance and posture from different subjects, which lays the foundation for cross-subject behaviour magnification and comparison. Fig. 2c presents the resulting synthesis when combining the appearance of row $i$ with the posture from column $j$ to generate $x^\prime_{ij} = D(E_\alpha(x_i),E_\pi(x_j))$. The results in Fig. 2c show that our model can transfer appearance from a subject to another, generating clean and realistic images. This is possible only if appearance and posture are correctly disentangled. 
In Extended Data Figure 6, we compare our method against a baseline approach, showing that disentanglement is a complex task, especially when posture annotations are not available.

\textit{Magnification Process.}
Using our disentanglement process, we can now magnify characteristic deviations in behaviour without altering the appearance. As visually depicted at the bottom of Fig. 1c, we magnify the deviation between a posture $z^{x_i}_\pi = E_\pi(x_i)$ of a query frame $x_i$ and a reference posture $z_\pi^{x_{NN}}$ using linear extrapolation (black line in Fig. 1c connecting $z_\pi^{x_{NN}}$, $z_\pi^{x_i}$ and $z^{x_i^\prime}_\pi$) to obtain the magnification $z^{x_i^\prime}_\pi$,
\begin{align}
    z^{x_i^\prime}_\pi &= z_\pi^{x_{NN}} + \lambda \cdot(z^{x_i}_\pi - z_\pi^{x_{NN}}),
\end{align}
where $\lambda$ is a predefined parameter. 
For high $\lambda$ (e.g. $\lambda > 4$), the network starts to produce unrealistic images, since this extrapolation is leaving the space of realistic postures. Thus, there is no risk of introducing new behaviour because the outputs for too large $\lambda$ are obviously unrealistic images. On the other hand, too small $\lambda$ (e.g. $\lambda < 2.0$) will not amplify the deviation enough to be useful for the user.
We found $\lambda = 2.5$ to be a good compromise in all our experiments.
$z_\pi^{x_{NN}} = \frac{1}{K} \sum^K_{j=1} E_\pi(x_j)$, the average posture of reference (e.g. healthy) behaviour is computed as the mean over the $K$ nearest neighbours of $x_i$ in the posture encoding space given all reference frames. Using the nearest neighbours guarantees that the reference posture is related to the query posture, e.g., both arise from a left step or both from a right step. The magnified image $x_{i}^{\prime} = D(z^{x_{i}^\prime}_\pi,z^{x_i}_\alpha)$ (bottom right of Fig. 1c) is produced by the decoder $D$ using the magnified posture $z^{x_i^\prime}_\pi$ and the appearance encoding $z^{x_i}_\alpha = E_\alpha(x_i)$.
The magnified image $x_{i}^\prime$ and reference $x_{NN}$ differ in appearance, since they show different subjects. To facilitate the visual comparison, we re-synthesise a reference $x^\prime_{NN}$ with the appearance $z_\alpha^{x_i}$ of the impaired subject and the posture $z_\pi^{x_{NN}}$ of the reference (as in Fig. 2c).
Fig. 6 shows several examples comparing humans (b) and rats (c) to their healthy references. The deviation between the impaired posture (second row) and the healthy reference (first row) is magnified (third row) for each frame. The differences to the healthy reference is measured (fourth row) and localised per pixel by highlighting the magnitude of deviation (fifth row) and its direction (sixth row). In Extended Data Figure 3, we measure the differences between an original frame and its magnification. The differences are consistently larger for impaired patients than for healthy subjects (two sided t-test, p=$2\times10^{-8}$). For comparison, we perform the same evaluation using the motion magnification method of Oh et al.\cite{deepmag}. The p-values of $0.13$ indicates that this method magnifies healthy and impaired behaviour indiscriminately.

Intuitively, the reader could think that performing any of the previous experiments on the magnified sequences should provide even better healthy-impaired comparison. However, this is practically not the case since the small deviations are already detected by our behaviour representation which is sensitive to fine-grained postures changes. Therefore, using magnified sequences for our behaviour analysis procedure produces similar results as without.

\subsection{Comparison with Key-point Based Approaches.}
The evaluation in Fig. 2e compares our learned behaviour representation against two established methods on behaviour analysis. 

Given the object location, JAABA extracts hand-crafted features from the video, such as locomotion and appearance. DeepLabCut (DLC) extracts keypoints after being trained on 1500 manually annotated frames of our rat dataset, which were randomly drawn across subjects pre and post stroke. 
By contrast, our approach has learned a non-parametric representation to measure similarity of motor behaviour. We then obtain an estimate of impairment of a query by comparing its similarity to healthy and impaired sequences. Each approach is trained to distinguish healthy and impaired sequences using the same training data and evaluated on held out test data. 
Extended Data Figure 7 shows how the number of annotated frames influences the performance of DLC. The features are based on 14 keypoints (see Supplementary). 
In Extended Data Figure 9, as complementary experiment, we compare DLC and our representation on the keypoint detection task.

Per day of rehabilitation, each grasping sequence needs to be classified as healthy or impaired by a classifier which is trained for the respective method. Since we do not have labels for each grasping sequence, but only for the entire video, this is considered a "weak" supervision in machine learning. Moreover, we do not know the level of impairment of each animal a priori. 
In Fig. 2e we utilise the Whishaw skilled reaching rating scale \cite{groundtruth} which is a standard method to manually quantify grasping behaviour. This manual annotation technique scores several movement types of the paw (for example, pronation and supination) during the grasp, providing a complete, but tedious, analysis of grasping behaviour. Fig. 2e compares our automatic scoring with the Whishaw score by means of correlation and Euclidean distance between the two scores. In parallel, we also compare DLC and Jaaba against the same manually annotated Wishaw ground-truth scoring.
The training set is composed of $1000$ sequences from pre-stroke videos (healthy) and $1000$ from two days post-stroke videos (impaired). The trained classifiers then predict the healthiness for each grasping sequence on a separate test set of $\sim90000$ sequences. For each day, we computed the relative number of sequences predicted as healthy. Fig. 2e presents the difference between this predicted frequency and the ground-truth frequency of healthy grasps per day. The correlation of our model with the ground-truth is $0.933\pm 0.005$, whereas Jaaba and DeepLabCut achieve $0.820\pm0.072$ and $0.695\pm0.014$, respectively. 
Similarly to our method, JAABA uses appearance features, which boost the performance compared to a keypoints-only approach. However, JAABA features and classification are based on classical, shallow approaches, whereas our representation is learned end to end. \\
More information on this experiment are provided in Supplementary, "Protocol for evaluation and comparison with DLC and Jaaba".

\subsection{Behaviour Analysis.}
In this section, we describe the experiment in Fig. 3 and 4.

\textit{Unsupervised Classification of Impairment based on Behaviour.}
We extracted the embedding $F_\beta(x)$ for each sequence $x$ in the mice on wheel dataset. Each high-dimensional embedding ($256$ dimensions) was projected into a 2D plot (Fig. 3a) using tSNE\cite{tsne} for dimensionality reduction, which retains the pair-wise distances between sequences. For pre- and post-stroke recordings, kernel density estimation (scikit learn implementation\cite{sklearn}) produces the two separate distributions shown in Fig. 3a. Being trained without supervision, the embedding $F_\beta(x)$ still captures characteristic features to provide a favourable separation. A linear classifier achieves an accuracy of $88\pm11\%$ on left-out test data confirming the visible separation in Fig. 3a.
The analysis of the mice dataset is limited to a binary classification because the mice subjects did not undergo a rehabilitation process (only pre and post
stroke recordings are available). Our subsequent evaluation focused on rats and humans, where the data for more fine-grained tasks is available.

\textit{Measuring the Similarity of Motor Behaviour.}
For all the experiments involving behaviour similarity shown in Fig. 3 and 4 we utilised a linear classifier $f(F_\beta(x)) = sign(W^T F_\beta(x))$ to distinguish the impaired sequences from healthy. The parameters $W$ of the classifiers are optimised using Linear Discriminative Analysis (LDA) \cite{fisher-lda}
(scikit-learn\cite{sklearn} implementation). LDA is, in contrast to support vector machines, more suitable for measuring similarities due to its generative properties.
During inference, we computed the classifier score $s = W^T F_\beta(x)$ as a measure of impairment or healthiness for every recorded sequence $x$, excluding the training set. 
We then compute a single score per video by averaging the scores over all sequences of that video.
We then computed the behaviour similarity between a reference (e.g. before stroke) and a query (e.g. 7 days after stroke) behaviour by calculating the overlap of the score distributions given all videos of the corresponding behaviours. Fig. 3 and 4 report the similarity of query to reference behaviour as the percentage of overlap between their score distributions.
For clarification, let us assume scores close to 1 indicating healthy and close to 0 impaired. The scores of the sequences before stroke are expected to be close to 1 and of behaviour 7 days after a stroke close to 0 apart from some outliers. Therefore, the two score distributions are unlikely to overlap, thus indicating a high dissimilarity in behaviour.

Fig. 4a and 4b show the similarity of motor behaviour on the rat data. In Fig. 4a we compare every day of training (-31d until -1d) against acquired skilled behaviour (0d). The classifier is trained on -31d against 0d with $\sim1000$ sequences each. 
We infer the score $s$ for all sequences and normalise all scores between $0$ and $1$ (min-max normalisation). Then, in order to produce a final score per day of training, we average the scores of all sequences belonging to that day. These final scores are then shown in Fig. 4a-top.
For analysing rat rehabilitation (Fig. 4b) we trained a classifier to compare the recordings during recovery against pre-stroke (0d) and post-stroke (2d) behaviour using $\sim1000$ sequences from pre-stroke and $\sim1000$ from post-stroke for training the classifier. 

Fig. 3b shows for the human data the distribution of scores of two classifiers: multiple sclerosis versus healthy and hydrocephalus versus healthy using for both $\sim1000$ sequences per class. For evaluating the effect of a treatment on human patients (Fig. 4c) we compare for both diseases post-treatment behaviour against healthy (vertical axis) and against pre-treatment behaviour (horizontal axis), respectively. The classifier is trained using sequences from healthy and pre-treatment.

\textit{Learning Skilled Motor Function Changes the Distribution of Postures.}
Fig. 4a bottom shows the changing frequency of different postures during the initial training (31days) of rats. After computing the posture embedding $F_\pi(x_i)$ for each frame $x_i$ in the dataset, the extracted postures are automatically sorted by projecting to one dimension using tSNE\cite{tsne} (vertical axis in Figure 4a, bottom). Grasping postures (middle) are separated from static postures (bottom). Kernel density estimation then yields the relative frequency of posture occurrences for each training day. Their difference from the distribution of skilled animals (0d) is displayed in colour.
\subsection{Relation between Brain Function and Behaviour.}
In this section we describe the method related to Fig. 5.

\textit{Neuronal Rewiring.}
Fig. 5a evaluates the recovery of motor-function during rehabilitation (vertical axis) and relates it to the neuronal rewiring (horizontal axis). The brain recovery is calculated by manually counting the number of BDA positive fibers post-mortem in the hemisphere affected by the stroke. For each animal, 10 brain slices were processed within 10 distinct positions with a focus on the sensorimotor cortex (from 4.68mm anterior to -5.04mm posterior to bregma with $\sim$1mm distance between slices) according to the Paxinos Rat Brain Atlas. The manual count of BDA positive fibers was determined on three consecutive slices per animal where the middle slice is selected to have the largest stroke area. To correct for variations in BDA labelling, data was normalised to the number of BDA-labelled axons in the intact corticospinal tract (CST) as described in \cite{wahl2017optogenetically}. As discussed in the Results section of the main manuscript, the outcome show that our behavioural score and the fiber counting have a satisfactory correlation of $\sim0.70$.

\textit{Optogenetic Stimulation.}
Fig. 5b shows the test accuracy of a linear support vector machine (SVM) classifier\cite{svm} when predicting whether the optical stimulation is activated during a video sequence. We split the recordings in train and test data and use the activation of the stimulus as label to train a classifier per animal and laser position using our behaviour representation $F_\beta$ as input. The test accuracy is then averaged across animals (Fig. 5b).
The classifier achieves random accuracy for control group, while it reaches almost $70\%$ for the treatment group. The results are discussed in the main manuscript in detail.

\section*{Experimental Setup and Data Acquisition.}
Information concerning the experimental setup for the optogenetic experiments in rats, single pellet grasping, BDA positive fibers as markers for neuronal plasticity, the running wheel task in mice and the acquisition of data in patients with neurological disease can be found in the supplementary.

\section*{Blinding.} 
For the behaviour analysis using uBAM investigators were blinded for the different treatment groups in both – human and rodent data. The biologists/clinicians only provided the code for the different patients and rodent data after completion of the uBAM behaviour analysis.

\section*{Statistics.}
Mean and standard deviation have been reported for experiments involving training a classifier. The classifiers have been trained several times by randomly splitting subjects into train/test set. The Pearson correlation coefficient in Figure 5a has been computed over each pair [fiber count, fitness score] for all animal subjects. In the subsection ''Behaviour Magnification'' of Results, we compute the two sided t-test for our method and Oh et al.\cite{deepmag} by testing the null-hypothesis whether the distributions of healthy and impaired deviation scores are the same.

\section*{Acknowledgements}
This work has been supported in part by the German Research Foundation (DFG) projects 371923335 and 421703927 to B.O. as well as the Branco Weiss Fellowship Society in Science and the Swiss National Foundation Grant (Nr. 192678) to ASW.

\subsection{Author contributions}
B.B., U.B. and B.O. developed uBAM. B.B. and U.B. implemented and evaluated the framework and M.D. and P.R. the VAE. A.S.W., L.F., and F.H. conducted the biomedical experiments and validated the results. B.B., U.B., and B.O. prepared the figures with input from A.S.W. and all authors contributed to writing the manuscript.

\subsection{Data Availability.}
The rats data can be downloaded at \url{https://hci.iwr.uni-heidelberg.de/compvis_files/Rats.zip}. \\
The Optogenetics data can be downloaded at \url{https://hci.iwr.uni-heidelberg.de/compvis_files/Optogenetics.zip}. \\
The mice data can be downloaded at \url{https://hci.iwr.uni-heidelberg.de/compvis_files/Mice.zip}.\\
The human dataset cannot be publicly released due to privacy issue. Please, contact the authors if needed.

\subsection{Code Availability.}
The code for training and evaluating our models is publicly available on github at the following address: \url{https://github.com/utabuechler/uBAM} (DOI: \cite{ourCode}).

\section*{COMPETING INTERESTS}
The authors declare no competing interests.

%
%

\section{Extended Data Figure 1}
\textbf{Qualitative comparison with the state-of-the-art in motion magnification}. To compare our results with Oh et al.(30), we show five clips from different impaired subjects before and after magnification for both methods. First, we re-synthesize the healthy reference behavior to change the appearance to that of the impaired subject so differences in posture can be studied directly, first row (see Method). The second row is the query impaired sequence. Third and forth rows show the magnified frame using the method by Oh et al.(30) and our approach, respectively. The magnified results, represented by magenta markers, show that Oh et al. corrupts the subject appearance, while our method emphasises the differences in posture without altering the appearance. (Details in Supplementary)

\begin{figure}
\centering
\includegraphics[width=0.99\linewidth]{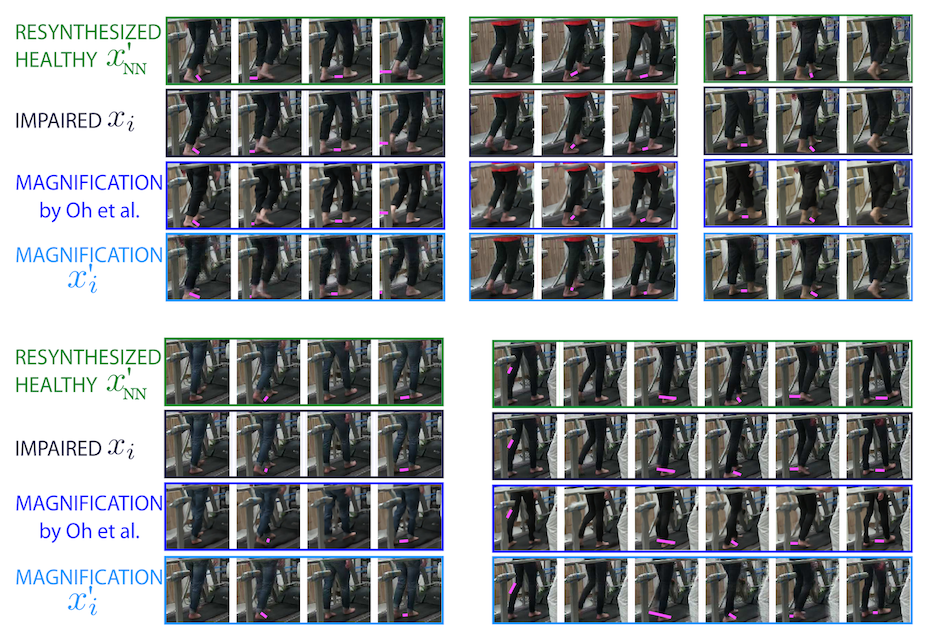}
\end{figure}

\section{Extended Data Figure 2}
\textbf{Quantitative comparison with the state-of-the-art in motion magnification}. a: mean-squared difference ($\text{white}=0$) between the original query frame and its magnification using our method and the approach proposed by Oh et al.(30). For impaired subjects, our method modifies only the leg posture, while healthy subjects are not altered. Oh et al.(30) mostly changes the background and alters impaired and healthy indiscriminately. b: Measuring the fraction of frames with important deviation from healthy reference behaviour for each subject and video sequence and plotting the distribution of these scores. c, mean and standard deviation of deviation scores per cohort and approach. (Details in Supplementary)

\begin{figure}
\centering
\includegraphics[width=0.99\linewidth]{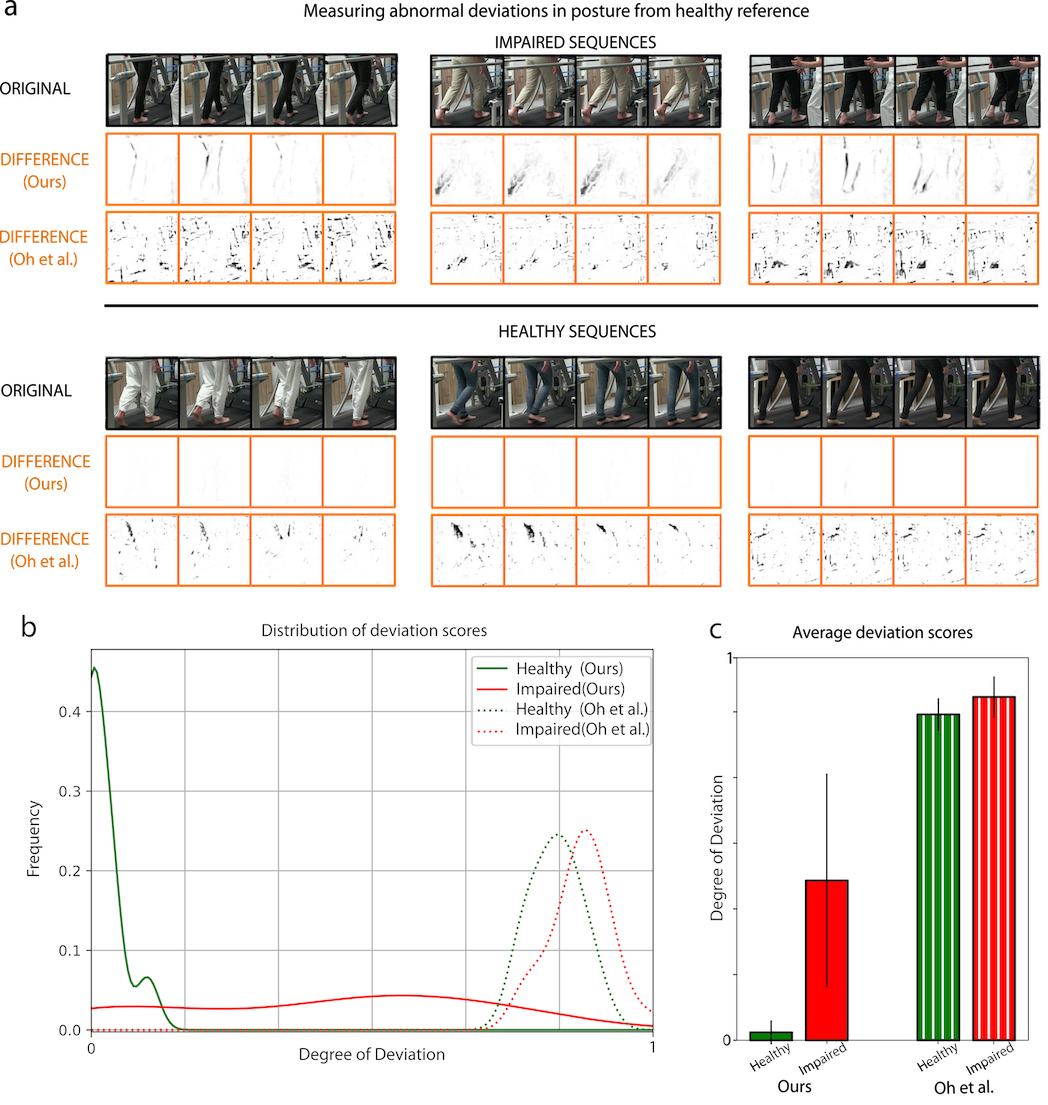}
\end{figure}

\section{Extended Data Figure 3}
\textbf{Abnormality posture before and after magnification}. We show that our magnification supports spotting abnormal postures by applying a generic classifier on our behaviour magnified frames. This doubles the amount of detected abnormal postures without introducing a significant number of false positives. In particular, we use a one-class linear-svm on ImageNet features trained only on one group (i.e. healthy) and predict abnormalities on healthy and impaired before and after magnification. The ratio of abnormalities is unaltered within the healthy cohort ($\sim2\%$) while it doubles in the impaired cohort ($5.7\%$ to $11.7\%$) showing that our magnification method can detect and magnify small deviations, but that it does not artificially introduce abnormalities. (Details in Supplementary)

\begin{figure}
\centering
\includegraphics[width=0.99\linewidth]{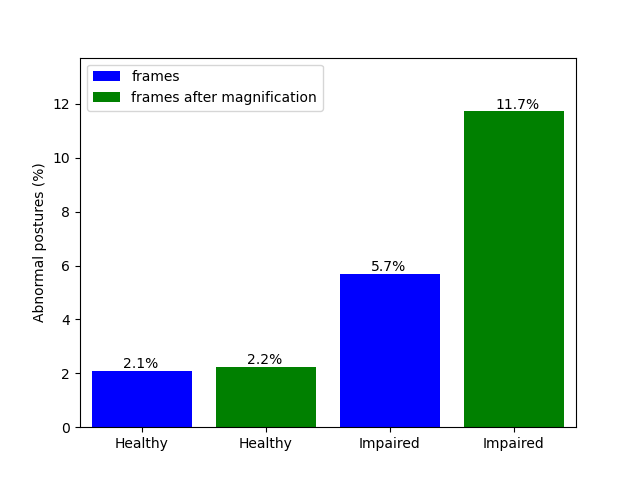}
\end{figure}

\section{Extended Data Figure 4}
\textbf{Qualitative evaluation of our posture encoding on the rat grasping dataset.} Projection from our posture encoding to a 2D embedding of $1000$ randomly chosen postures using tSNE. Similar postures are located close to each other and the grasping action can be reconstructed by following the circle clockwise (best viewed by zooming in on the digital version of this figure). (Details in Supplementary)

\begin{figure}
\centering
\includegraphics[width=0.99\linewidth]{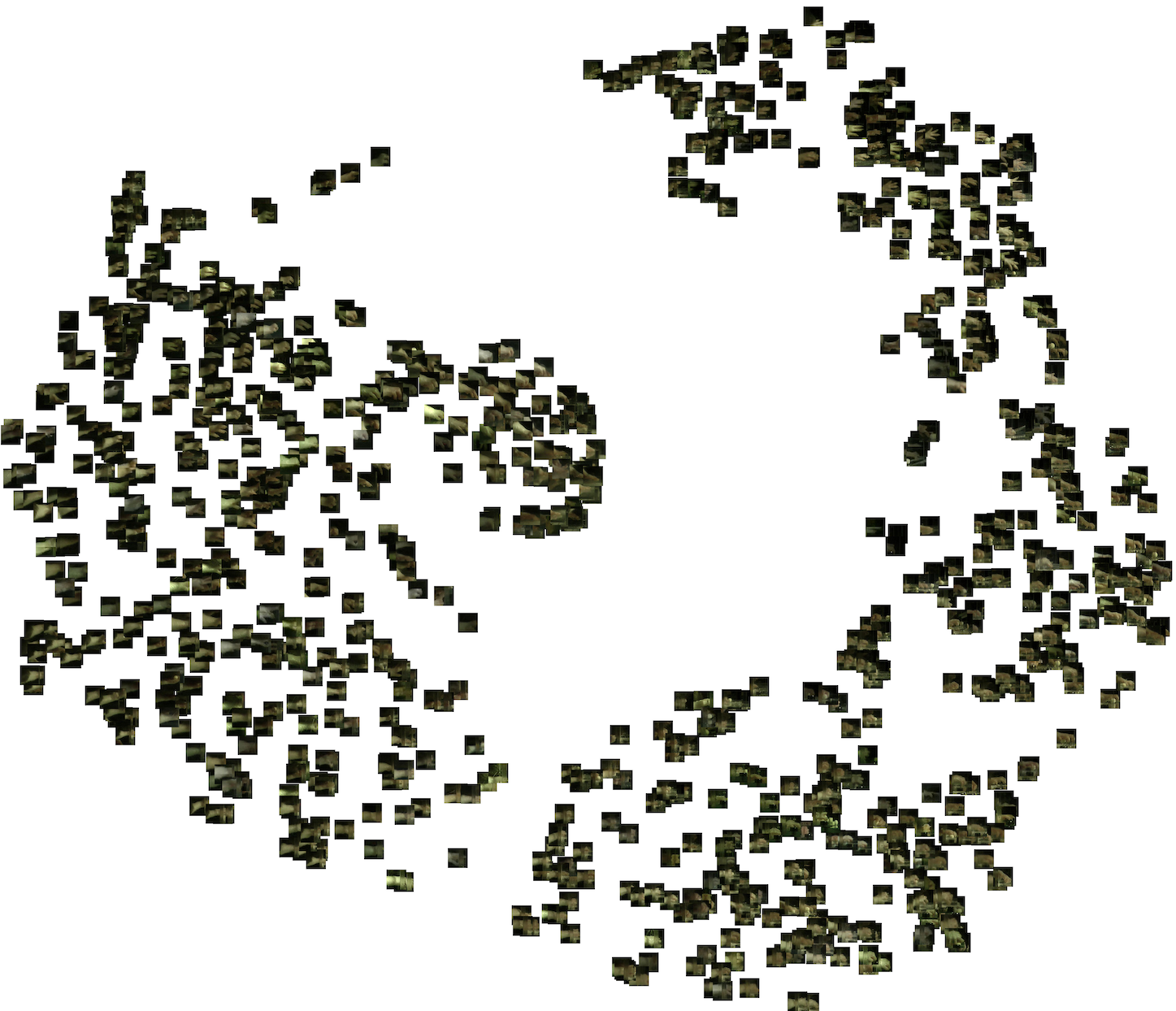}
\end{figure}

\section{Extended Data Figure 5}
\textbf{Comparison with PCA of posture encoding}. \textit{a}: A single video clip projected onto the two most important factors of variation using PCA directly on RGB input (left) and our representation (right). Consecutive frames are connected by straight lines colourised according to the time within the video. Every four frames we plot the original frame. PCA is able to sort the frames over time automatically, showing that each cycle is overlapping with the previous one. Our representation better separates different postures thus reflected by the circular shape of the embedding. \textit{b}: same as \textit{a} but including more videos. Each colour represent a different subject. In this case, PCA is strongly biased towards the subject appearance. Thus it separates subjects and does not allow to compare behaviour. \textit{c}: We reduce the appearance bias by normalising per video with the mean appearance. The result still shows subject separation and no similarity of posture across subjects. \textit{d}: Using our posture representation and applying PCA on $E_\pi$ instead of directly on video frames shows no subject bias and only similar postures are near in the 2D space. (Details in Supplementary)

\begin{figure}
\centering
\includegraphics[width=0.99\linewidth]{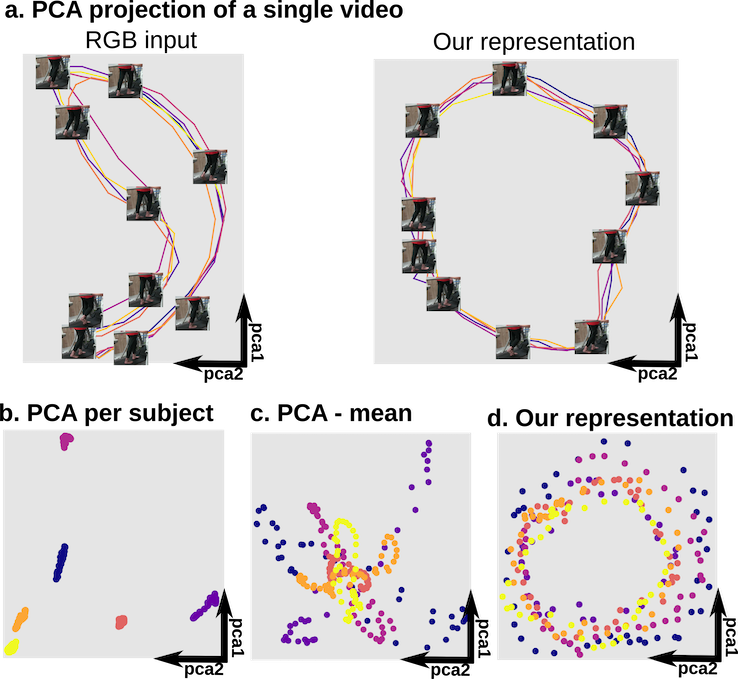}
\end{figure}

\section{Extended Data Figure 6}
\textbf{Disentanglement comparison with simple baseline}. We transfer posture from a subject (row) to others with different appearance (columns). \textit{a}: A baseline model which uses the average video frames as appearance. The appearance is subtracted from each frame to extract the posture. \textit{b}: Disentanglement using our custom VAE for extracting posture and appearance. Checking for consistency in posture along a row and for similarity in appearance along a column shows that disentanglement is a hard problem: a pixel-based representation cannot solve the task, while our model produces more detailed and realistic images. (Details in Supplementary)

\begin{figure}
\centering
\includegraphics[width=0.99\linewidth]{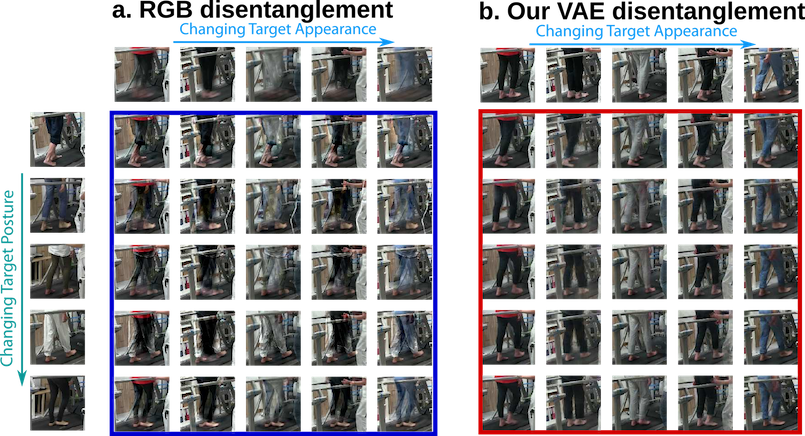}
\end{figure}

\section{Extended Data Figure 7}
\textbf{DeepLabCut trainset size}. We train DLC models on a growing number of training samples. The model is evaluated as described in Fig. 2 of the main manuscript. Note the limited gain in performance despite annotation increasing by more than an order of magnitude. (Details in Supplementary)

\begin{figure}
\centering
\includegraphics[width=0.99\linewidth]{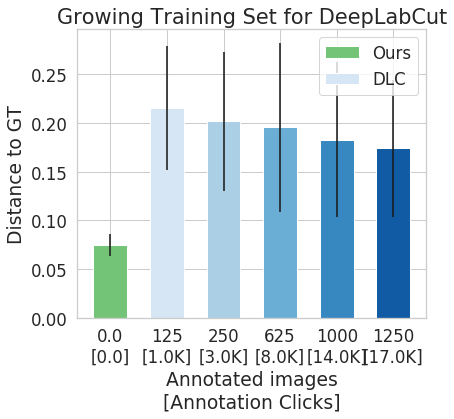}
\end{figure}

\section{Extended Data Figure 8}
\textbf{Comparison with R3D}. Besides JAABA and DLC we also compare our method with R3D which is another non-parametric model, very popular for video classification. We extract R3D features and evaluate the representation using the same protocol as our method. Our model is more suited to behaviour analysis. More information regarding the evaluation protocol can be found in the Methods section of the main manuscript. (Details in Supplementary)

\begin{figure}
\centering
\includegraphics[width=0.99\linewidth]{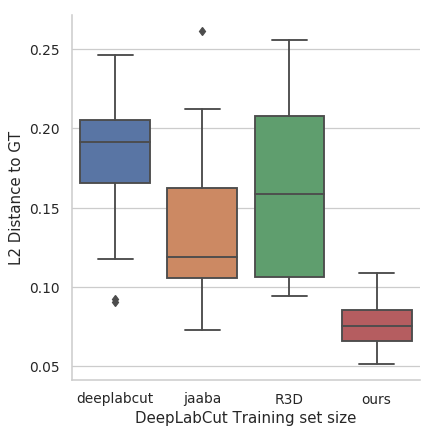}
\end{figure}

\section{Extended Data Figure 9}
\textbf{Regress Key-points}. We show qualitative results for the key-point regression from our posture representation to key-points and ene-to-end inferred key-points for DLC. This experiment was computed on 14 keypoints, however we only show 6 for clarity: wrist (yellow), start of the first finger (purple), tip of each finger. The ground-truth location is shown with a circle and the detection inferred by the model with a cross. Even though our representation was not trained on keypoint detection, for some frames we can recover keypoints as good as, or even better, than DLC which was trained end-to-end on the task. We study the gap in performance in more detail in the Supplementary (Supplementary Figure 3).

\begin{figure}
\centering
\includegraphics[width=0.99\linewidth]{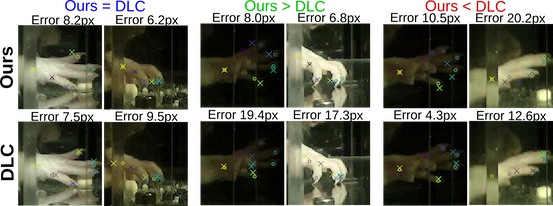}
\end{figure}

\section{Extended Data Figure 10}
\textbf{Typical high/low scoring grasps with optogenetics}. Given the classifier that produced Figure 5b, we score all testing sequences from the same animal and show two typical sequences with high/low classification scores. The positive score indicates that the sequence was predicted as light-on, the negative that it was predicted as light-off. Both sequences are correctly classified as indicated by the ground-truth ("GT") and classifier score ("SVM-Score"). The sequence on the left shows a missed grasp, consistent with a light-on inhibitory behaviour, while the same animal performs a successful grasp in the sequence on the right for the light-off. Obviously, the classifier cannot see the fiber optics, since we cropped this area out before passing it to the classifier. (Details in Supplementary)

\begin{figure}
\centering
\includegraphics[width=0.70\linewidth]{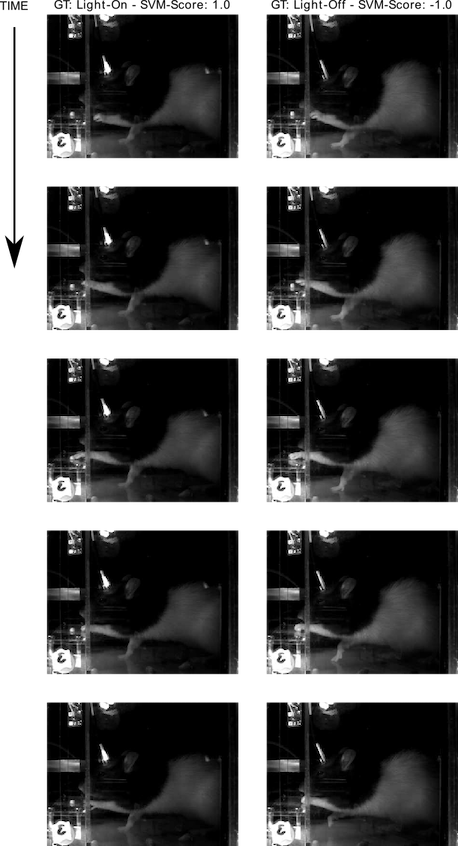}
\end{figure}

%
%
%
%
\clearpage
\section*{REFERENCES}
\vspace{15px}
\bibliography{ms}{}

\bibliographystyle{naturemag}
\end{document}


\begin{center}
{\LARGE \textbf{Supplementary Information}}\\
\end{center}

\section{ADDITIONAL METHODS}
\subsection{Pipeline}
Here, we provide a brief step-by-step overview of the processing pipeline which is also visually illustrated in Fig. 1a. The details of our architecture, the training procedure, and the experimental setup follow in later sections.
\\ \\
\textit{Pre-processing.}
Even though we focus on constrained laboratory settings, we still need to indicate which subject or region to focus on in each video. For example, in the human gait dataset a nurse is often standing and moving next to the patient. Thus, for each video, a user once has to provide one bounding-box pointing at the desired subject. This is sufficient due to the static camera setup and the on the spot movement of the patient on the treadmill. The same procedure can be applied for the mouse dataset. In contrast for datasets with significant overall movement of the subject through the scene, such as the paw movement in the rat stroke dataset, we need a tracking mechanism which discovers the subject in every frame of a video in order to capture the fine-grained details of the behaviour.  
For the rat stroke dataset we apply optical flow and robust PCA \cite{wright2009robust} (for separating the foreground movement from the static background) to discover the moving paw and tracking it through the entire video. To restrict the analysis to only a specific paw, the user can also provide a single initial bounding box for only the first video frame which is then automatically tracked. 
\\ \\
\textit{Sequence Sampling.}
The sequence length is proportional to the periodicity of the motion; it is a rough estimate of the number of frames which represent a full motion: human gait: 27 frames ($\sim 1 s$), rat grasp: 8 frames ($\sim 0.3 s $) and mice step: 20 frames ($\sim 0.4 s $). Substantially shorter sequences will result in an incomplete motion and longer sequences are not recommended since multiple motions will be included in a single sequence. Nevertheless, the model is robust to small variations: we tried up to 32 frames for the human dataset and up to 11 for the rats without obtaining a significant change in the results. For our experiments, we then densely sampled the sequences using a step of 1 and the lengths given above.
\\ \\
\textit{Training of Posture and Behaviour Representation.}
In order to analyse the behaviour of subjects during rehabilitation without any human annotations we require an unsupervised model that is able to encode the video data into a lower dimensional representation which focuses solely on the information necessary to compare behaviour. A classical approach for removing data variability by projecting into a lower dimensional space is Principle Components Analysis (PCA) (see Extended Data Figure 4). Additionally, Independent Component Analysis (ICA) can disentangle the input in different components. However, those approaches offer little control over what information is preserved by the lower dimensional vector. Hence we cannot compare across individuals since there is no guarantee of separation of posture from appearance.
After extracting the bounding box of the object-of-interest for every frame in the pre-processing step, we randomly collect several overlapping sub-sequences from every video in order to create a large set of training samples.
Then, we train a neural network on the collected training samples using our unsupervised training protocol which is illustrated in Fig. 1b and described in detail in the subsequent paragraph \textit{Unsupervised learning of the representation}. In short, we train the neural network on a surrogate task which does not require any manual labels but enables the network to learn meaningful posture and behaviour features directly from the data. Once the deep neural network is trained, we can use it to encode our video data. In particular, we extract two types of information from our trained network: \textit{(i)} a behaviour representation $F_\beta$ to objectively quantify the changes in behaviour during rehabilitation (as illustrated in Figure 1a (middle)) and \textit{(ii)} posture features $F_\pi$ for training our generative model which will be described subsequently.
\\ \\
\textit{Generative Model Training.}
To provide a complete diagnostic support system, we aim to discover and visualise behavioural deviations of impaired subjects by comparing them with healthy behaviour. Thus, we propose a generative model for magnifying small deviations in posture between a query and reference subject. For that matter, we require the model to magnify solely posture deviations while keeping the appearance unaltered. Thanks to the previous training step, we already obtained a fine-grained posture representation $F_\pi$ which does not contain any appearance information. We employ the posture representations extracted in the previous step and an additional objective function to train our model on extracting the appearance, while keeping it disentangled from posture, and image generation (more details see \textit{Behaviour Magnification by Disentangling Posture and Appearance}). During inference, we can now use the trained generative model to \textit{(i)} separately encode posture $E_\pi$ and appearance $E_\alpha$ of the query and reference subject \textit{(ii)} magnify their deviations solely in the posture encoding and \textit{(iii)} decode the resulting magnified posture representation back to the pixel space using the appearance of the query subject (see Fig. 1c (bottom)). The output magnifications simplify the discovery of behavioural deviations by a user, e.g. a clinician, and the localisation of their origin and consequent interpretation of symptoms.
\\ \\
Practically it is composed of five convolutional layers (\textit{conv1} to \textit{conv5}) with intermediate max-pooling, and two fully connected (\textit{fc6} and \textit{fc7}), including standard ReLU activation between each layer. 
To incorporate temporal information, the output produced by the convolutional neural network\cite{alexnet} (CNN) for each frame $x_i$ of a sequence $x$ is concatenated using a recurrent layer, i.e., a long short-term memory layer\cite{lstm} (LSTM). For our experiments, we utilised an hidden state of 1024 nodes. 
During inference, the \textit{fc6} layer is used as posture encoder $F_\pi$ and the LSTM hidden state is the behaviour encoder $F_\beta$. Finally, a fully connected layer, acting as classifier, predicts if the input sequence $x$ is real or permuted. Based on the binary cross-entropy loss the network parameters are optimised using stochastic gradient descend (SGD) with weight decay. We use a learning rate of $0.01$ and batch size of $48$ sequences ($24$ real and $24$ permuted). Each image $x_i$ is resized to $227 \times 227$. A network is trained for each dataset: The representation is trained using $\sim100000$ sequences from the rat grasping videos with minimum length of $8$ frames per sequence, $\sim170000$ sequences with minimum length of $27$ frames for humans, and $\sim18000$ with a length of $20$ frames for the mice. For all our experiments, we implemented the network in PyTorch and utilised a single NVIDIA TitanX for training.
\\ \\
\textit{Summary of the Used Similarity Measures.} Several similarity measures can be computed on $F_\pi(x)$ and $F_\beta(x)$ in order to analyse the behaviour. In our specific case, we use the cosine similarity to query nearest neighbours in both posture and behaviour space and we compare behaviours of different animal cohorts by computing the overlap of their $F_\beta$ distributions between different groups (e.g. pre and post treatment, beginning and end of training and rehabilitation). Another similarity is introduced in the generative model section, which compares images based on VGG features. This is used in the perceptual loss \cite{perceptualLoss} during training.
\\ \\
\textit{Searching Similar Postures and Behaviour without Annotations.} Learning an encoding means understanding which samples are similar based on particular features (posture or behaviour) while ignoring others (appearance). 
To demonstrate which characteristics are learned by the network and which are ignored, we computed the cosine similarities between all samples of the dataset in the representation space $E(x)$ and show nearest neighbours in Fig. 2a,b. The visualisation shows that image similarity is based on the posture while paw appearance (white or coloured skin) is having no influence, hence our representation contains posture information and no appearance.
(Fig. 2a,b, Extended Data Figure 1 and Supplementary Figure 1). This also demonstrates the suitability of our encoding for searching similar postures, and sequences, without the need for manually annotated data.
\subsection{Protocol for Evaluation and Comparison with DLC and Jaaba.}
Given the diverse characteristics of the three compared approaches, we adapted the evaluation process accordingly to achieve the best possible outcome for each approach. \\
Jaaba requires the user to manually annotate the training frames by clicking on the object location within the software. 
Therefore, we bypassed the manual labelling part and provide the same input used by our approach. Then we continued with the normal workflow of the software: selected all available features and a time window of 8 frames (same as our method); trained the model to classify healthy/impaired and inferring the remaining grasps. Finally, we compare the classification scores against the other methods. \\
As a second method for comparison, we use DeepLabCut (DLC). DLC provides a key-point detector network for human pose estimation pretrained on $\sim25000$ labelled images. A model is also available for rodents, however it does not fit our choice of keypoints and data format since the recordings are taken from the front and only 13 keypoints are detected. Therefore, for a fair comparison, we fine-tuned the network on our own data following the step-by-step instruction provided by the authors. We defined 14 key-points on the rat paws: 3 per finger (for 4 fingers), wrist and arm. We manually labelled 1500 frames randomly sampled from a diverse set of videos across pre and post stroke. 
In Extended Data Figure 6, we study the performance of DLC relatively to the number of train samples, showing that, in the best case scenario, we would need at least 5 times more data to reach performances similar to ours in behaviour analysis. However, this is costly in terms of human labor and time. Detecting the exact location of body-parts is a hard task due to variance in human annotations (e.g. different annotators disagree on where a key-point should be placed), occlusion and brightness/contrast changes. 
For the task of behaviour analysis we can circumvent these issues by learning a behaviour representation that directly maps the input video pixels onto a representation that is optimised for the final analysis task rather than having users specify an intermediate representation using keypoints. This loss of information explains why DLC performance shows to converge to a weaker outcome than that of our model in Extended Data Figure 6.
After all, end-to-end learning of representations which are directly tuned for the task at hand was key for the success of deep learning.
During inference, we applied the fine-tuned DLC model on all data to detect the 14 keypoints. Then, we extract features from the xy keypoint locations, their trajectories and velocity over 8 frames. These features have been tested with several classifiers to predict healthy and impaired: support vector machines, AdaBoost with Decision Trees, and a multi-layer perceptron (MLP). AdaBoost performed best for this final evaluation. AdaBoost combines several weak estimators to form a stronger one and cross-validation showed that Decision Trees worked best as estimator with a maximum depth of 4 and a total of 30 estimators. MLP with two layers was a close second, but we decided for the AdaBoost since MLP tends to overfit when few data are available.\\
To evaluate our model we trained a linear support vector machine\cite{svm} using our behaviour embedding $F_\beta(x)$ as input. After optimising the hyper-parameters by cross-validation, we evaluated the model using a linear kernel and a soft-margin value of 2.0. The classifiers and the cross-validation have been implemented using scikit-learn\cite{sklearn}.
\subsection{Encoder-Decoder Architecture for Behaviour Disentanglement.}
The variational encoder-decoder architecture is composed of three networks: the posture encoder $E_\pi$, the appearance encoder $E_\alpha$, and the decoder $D$. $E_\pi$ is the same network described in section 'Architecture for behaviour representation'.
$E_\alpha$ has the same structure as $E_\pi$, so it is composed of 5 layers, each layer includes a convolutional layer with 4 kernels each and stride 2.
On top of both encoders, we introduce a batch normalisation and LeakyReLU which are typically beneficial for generative models.
For the variational part, we include on top of $E_\pi$ and $E_\alpha$ two fully connected layers, one for producing the mean and one for the standard deviation. 
The dimensionality of the encoding is $100$ ($50$ mean and $50$ std).
During inference it is standard to use directly the mean without sampling from the standard deviation.
Following common practices, the decoder has the same number of layers as the encoders.
Firstly a fully connected layer with ReLU receives the appearance and posture encoding as input (concatenated); the output is then reshaped and input into 5 transpose convolution layers. Each transpose convolution layer contains upsampling, a convolution operation with kernel size 3, batch normalisation, and LeakyReLU. We implement and train the network using the PyTorch deep learning framework. The parameters of the network are optimised using Adam solver, with a learning rate of $5\cdot 10^{-4}$, while keeping the default values for the other hyper-parameters. We train the model for 50 epochs on a single GPU (NVIDIA TitanX).
\subsection{Previous Work on Video Magnification.} 
Existing methods on video magnification has mainly addressed the magnification of small motions \cite{deepmag, magnification, wu2012eulerian, elgharib2015video, wadhwa2013phase, wadhwa2014riesz, zhang2017video, tulyakov2016self, revealingnonlocal} or the deviation from a predefined reference shape \cite{deviationmagn}, but only within the same video \cite{magnification,deepmag}.
Moreover, previous work has often been limited to repetitive motion \cite{wu2012eulerian, wadhwa2013phase, wadhwa2014riesz}, since these methods only work in the frequency domain.
While Balakrishnan et al. \cite{videodiff} has focused on discovering differences between videos based on object boundary contours, it only detects differences but does not magnify the deviations. Being only based on edge pixels, the approach can be mislead by spurious differences that are irrelevant to behaviour.
Oh et al.\cite{magnification} use a deep generative neural network trained on synthetic data for motion magnification. 
The approach can amplify motion within the video but not compare objects across videos. 
In Extended Data Figure 2 and 3 we compare this model with our method specifically for posture magnification. Their model is able to disentangle moving objects from static. However, our aim of disentangling posture and appearance is a more fine-grained task. In particular, our model needs to learn the commonalities of behaviour performed by different healthy reference subjects and disregard the rest.
\subsection{Experimental Setup and Data Acquisition.}
Subjects of this study were a total of n=36 adult female Long-Evans rats ($200-250$g, 3-4 months of age, Janvier, France) housed in groups of two to four. Moreover, experimental evaluation included n=9 adult transgenic ChR2 mice (Thy-1COP4/EYFP)\cite{arenkiel2007vivo}, aged 2-4 months, weighed $19-23$g of both sexes. Mice were housed in groups of two to six. All animals were kept under a constant 12h dark/light cycle with food and water ad libitum. All experimental procedures were approved by the veterinary office of the canton of Zurich, Switzerland.
\\ \\
\textit{Single Pellet Grasping Task in Rats.}
The experimental setup and the sequence of events for the treatments of the rats whose grasping behaviour is analysed here can be found in Wahl et al\cite{wahl2017optogenetically}. In brief, rats were trained in the single pellet grasping task as previously described\cite{wahl2017optogenetically,wahl2014asynchronous} to assess fine motor control of the forelimb: Animals were placed in a Plexiglas box (34 x 14 cm) with two openings on opposite ends and were trained to grasp pellets. To stimulate distinct corticospinal projecting fibers after stroke, a retrograde AAV9-CamKII0.4.Cre.SV40 vector (Penn Vector Core, Philadelphia) was injected in the contralateral cervical hemi spinal cord of the preferred paw (contralateral to the ‘future’ denervated cervical hemi spinal cord) followed by the injection of a Cre-recombinase dependent channelrhodopsin-2 vector (AAV2.1 Ef1a-DIO-hChR2(t159C)-mCherry, UNC, Chapel Hill) in the ipsilateral pre- and sensorimotor cortex (in the ‘future’ contralesional hemisphere) as previously described\cite{wahl2017optogenetically}. Three optical implants were then positioned over the pre- and sensorimotor cortex, where the viruses had been injected\cite{wahl2017optogenetically}. After recovery from the surgeries, rats were trained up to 5x/ week until they successfully gasped for at least 60\% of the provided pellets. All rats then received a photothrombotic stroke targeting the sensorimotor cortex corresponding to their paw-preference in the grasping task (contralateral to the hemisphere of the fiber implantation) as previously described\cite{wahl2017optogenetically,wahl2014asynchronous}.  Two days after stroke, animals were re-assessed in the single pellet grasping task and randomised in four different treatment groups according to their lesion deficit: Animals in the ‘Stimulation’ group and ‘Stimulation and Training’ group received optical stimulation of the intact corticospinal tract with blue light (473nm wave length, 3 x 1 minute stimulation with 10 Hz, 20 ms pulses) 3x/day from day 3 to day 14 after stroke. Additionally to the optical stimulation, animals in the ‘Stimulation and Training’ group underwent intensive grasping training of the impaired paw (grasping for at least 100 sugar pellets/training session) during the 3rd and 4th week after stroke. Animals in the ‘Delayed Training’ group were also intensively trained in the single pellet grasping task during the 3rd and 4th week after stroke but without optical stimulation in advance. In the ‘No treatment’ group, animals were only assessed for regain of grasping function weekly up to 4-5 weeks after stroke. All training and testing sessions were filmed (Panasonic HDC-SD800 High Definition Camcorder, 50 frames/s) for further analysis of grasping postures and kinematics.
\\ \\
\textit{BDA Positive Fibers as Measurement of Neuronal Rewiring in the Peri-infarct Region.}
6 weeks after stroke, after the completion of the training and testing in the single pellet grasping task, the intact contralesional motor cortex of rats was traced anterogradely with Biotinylated Dextran Amine (BDA, 10,000 molecular weight, 10\% solution in 0.01 M PBS, Invitrogen) as previously described\cite{wahl2014asynchronous}. Three weeks after BDA injections, animals were euthanized and the further histological work-up including the DAB staining was performed\cite{wahl2014asynchronous}. Brain sections were analysed for BDA-positive neurons in the stroke-damaged hemisphere as a parameter for neuronal rewiring taking place in the peri-infarct cortex.
\\ \\
\textit{Optogenetic Silencing of Corticospinal Circuitry during Single Pellet Grasping.}
We measured the effect that optogenetic silencing of subregions of the intact pyramidal tract has on the grasping ability of rats after stroke. Therefore, we expressed the light-sensitive inhibitory proton pump ArchT in a Cre-dependent approach as previously described\cite{wahl2017optogenetically}. Before virus injection, animals in the ‘Treatment’ group had received a growth-promoting immunotherapy followed by rehabilitative training after stroke as described in previous work\cite{wahl2014asynchronous}, while no treatment was applied to the ‘Control’ group. Up to 7 days after virus injection we implanted three custom-made optic glass fibers in the pre- and primary motor cortex\cite{wahl2017optogenetically} to enable the inhibition of corticospinal neurons originating from three distinct positions (position 1: pre- and rostral motor cortex; position 2: Motor cortex; position 3: motor cortex with small parts of the primary sensory cortex). For optogenetic silencing at these three positions, animals were put in a Plexiglas box with their optical implants connected via ceramtic mating sleeves to 1m long optical fibers ({\o}\SI{400}{\micro\metre} Core, 0.39 NA, Thorlabs) and three lasers (532nm) above the grasping box, enabling independent optical inhibition at each optical implant location. A light barrier regulated laser on- and off-set times as it was positioned at the centre of the provided food pellet so that by the first grasp through the light barrier a program (LabVIEW, National Instruments) was started which kept the lasers off for 100s and turned them on afterwards for 100s while the animal continuously grasped for pellets. Animals were either stimulated at one position or at all three positions at the same time during one grasping session. All training and testing sessions were filmed with  a Panasonic HDC-SD800 High Definition Camcorder, 50 frames/s.
\\ \\
\textit{Running Wheel Task for Mice.}
Mice were handled and habituated to locomote on top of a 23-cm diameter wheel with irregular rung distance (between 0.5-3cm) while being head-fixed\cite{pilz2016functional}. Mice received ‘baseline’ training for at least 5 days prior to a photothrombotic stroke surgery and were then re-assessed 1-3 days after stroke. Training and testing sessions were recorded at 90 Hz frame rate (1280x 640 pixels) using a high-speed CMOS camera (A504k; Basler).
\\ \\
\textit{Gait Analysis in Healthy Subjects and Patients with Neurological Disorders.}
14 patients diagnosed with relapsing-remitting, primary- or secondary-progressive MS as well as 18 patients with diagnosed hydrocephalus and 9 healthy controls were assessed at University Hospital Zurich between 2017 and 2018. The project was approved by the Zurich cantonal ethics committee (project-ID: 2017-01459, project title “An automatised computer-vision based algorithm to quantify sensorimotor deficits in animal models and patients with neurological disorders”) and was conducted in accordance with the Declaration of Helsinki and Good Clinical Practice. All patients had walking impairment. All participants walked barefoot on an instrumented treadmill. Locomotor assessments were performed while participants walked at fixed walking speeds of 1km/h, 2 km/h and 3km/h for 30 seconds each. Recordings were performed by a HD-camera (Canon, Legria HF G25; 25Hz sampling rate). The camera was attached onto a mobile stand (tripod). The height and position (marks on the floor) of the tripod was kept constant for all measurements. Patients with suspected hydrocephalus were assessed on the treadmill before and after lumbar puncture. We also analysed the effect of fampridin treatment on the walking abilities of MS patients by recording walking before and after fampridin therapy (for details of the fampridin treatment see Filli et al.\cite{filli2019predicting}).

\clearpage
\section{ADDITIONAL RESULTS}
In this section, we provide additional results in form of figures and tables to further extend our evaluations presented in the main manuscript.

\subsection{Posture and Behavior Encoding} Fig. 2a,b of the main manuscript presents Nearest Neighbours to qualitatively verify our posture and behaviour encoding. We now provide additional qualitative evaluations in Supplementary Figure 1 and Extended Data Figure 1.
In particular, we show projections of $1000$ randomly chosen postures from our posture encoding $E_\pi$ to a 2D embedding space using the dimensionality reduction approach \textit{tSNE}\cite{tsne}. As expected, it shows that similar postures are located close to each other (best viewed by zooming in on the digital version). For the human dataset (Supplementary Figure 1), it is possible to reconstruct the walking gate by following the circle anticlockwise, while for the rats (Extended Data Figure 1) the grasping action is automatically encoded clockwise, even though no human supervision has been used during the entire process from training the network until extracting the encoding for the 2D projection.
\\ \\
\textbf{Quantitative Evaluation of Posture and Behavior.} To provide a quantitative analysis without using keypoints, we manually select and provide 10 similar and 10 dissimilar queries for each reference. For the posture, we labelled 30 references and 22 for sequences. Then, we rank the 20 queries for each reference using cosine similarity on our representation. Finally, we calculate the accuracy as the ratio of similar queries within the first 10 nearest neighbor. Supplementary Tab.~1 shows the results for the posture similarity and Supplementary Tab.~2 for the sequence/behavior similarity.
Our posture representation was also tested\cite{lstm2017cvpr} on standard human posture benchmarks: Olympic sports\cite{olympic_sports} (Supplementary Tab.~3) and LEEDS\cite{leeds} (Supplementary Table~4).
The results show that our posture and behavior representation outperforms other non-parametric, unsupervised baselines by a large margin.

\begin{table}[t]
\centering
\begin{tabular}{|l|c|}
\hline
\textbf{Models}         & \textbf{Accuracy(\%)}\\
\hhline{|=|=|}
Imagenet \cite{alexnet}  & 65.3         \\
Our      & \textbf{85.6}    \\
\hline  
\end{tabular}
\caption{Evaluation of posture representation using the benchmark test set for posture similarity. \textit{Imagenet} refers to a fully supervised AlexNet network trained on the Imagenet dataset.}
\label{tab:tableFrameSim}
\end{table}

\begin{table}[t]
\centering
\begin{tabular}{|l|c|}
\hline
\textbf{Models}                    & \textbf{Accuracy(\%)} \\
\hhline{|=|=|}
Max frame similarity          & 74.1         \\
Avg frame similarity          & 75.9         \\
DTW\cite{DTW}		          & 76.8         \\
ClusterLSTM                   & 64.0         \\
Our        & \textbf{80.5}         \\
\hline
\end{tabular}
\caption{Evaluation of behavior representation using the benchmark test set for sequence similarity. The first two methods simply compute the similarity frame-by-frame and select the highest similarity (Max) or use the average. Dynamic Time Warping\cite{DTW} (DTW) was the standard for sequence similarity before recurrent neural networks became popular. ClusterLSTM trains the network using clustering as surrogate task. }
\label{tab:tableSeqSim}
\end{table}

\begin{table}[t!]
\small
\centering
\begin{tabular}{|l||c|c|c|c|c|c|}
\hline
Category
& HOG-LDA \cite{HOG-LDA}
& Ex.SVM \cite{ExemplarSVM}
& Ex.CNN \cite{ExemplarCNN}
& Alexnet \cite{alexnet}
& CliqueCNN \cite{CliqueCNN}
& Ours\\
\hline
Mean				& 0.58	& 0.67 & 0.56 & 0.65 & 0.79		& \textbf{0.83}\\
\hline
\end{tabular}
\caption{Average AUC of all categories of the Olympic Sports dataset using the state-of-the-art and our approach.}
\label{tab:OlympicSports}
\end{table}

\begin{table}[t!]
\small
\centering
\begin{tabular}{|l||c|c|c|c||c|c|c|}
\hline
Parts
& HOG-LDA \cite{HOG-LDA}
& Alexnet \cite{alexnet}
& CliqueCNN \cite{CliqueCNN}
& Ours
& PoseMach. \cite{PoseMachines}
& DeepCut \cite{deepcut}
& GT\\
\hline
Mean & 38.4 & 41.1 & 43.5 & \textbf{46.6} & 67.8 & 85.0 & 69.2 \\
\hline
\end{tabular}
\caption{PCP measure (observer-centric) of the Leeds Sport dataset using all mentioned approaches.}
\label{tab:Leeds}
\end{table}
\label{tableLeeds}

\begin{figure}
\includegraphics[width=\linewidth]{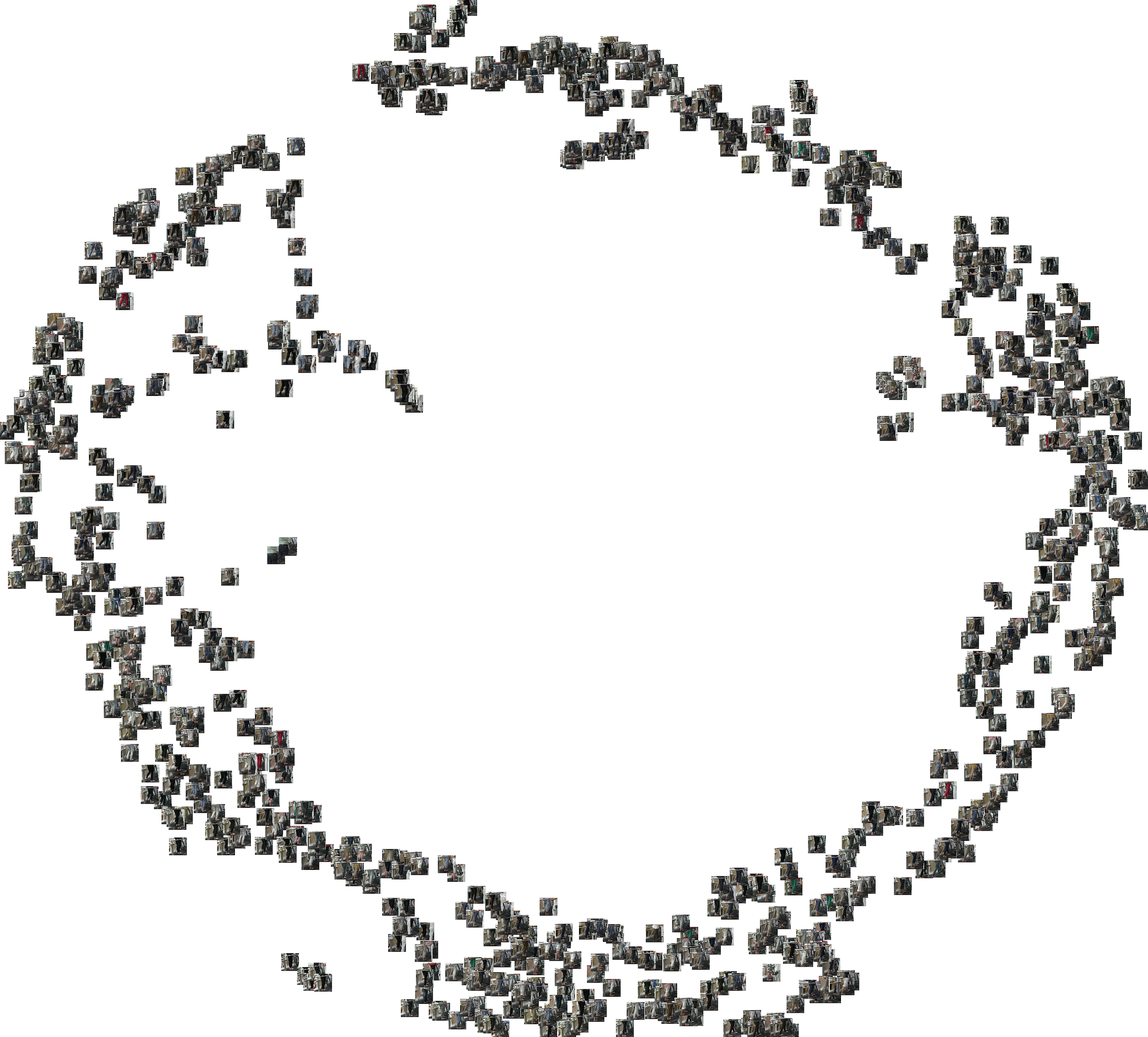}
\caption{\textbf{Qualitative evaluation of our posture encoding on the human gait dataset.} Projection from our posture encoding to a 2D embedding of $1000$ randomly chosen postures using tSNE. Similar postures are located close to each other and the walking gate can be reconstructed by following the circle anticlockwise (best viewed by zooming in on the digital version of this figure).}
\end{figure}

\begin{figure}[h!]
    \centering
    \includegraphics[width=\linewidth]{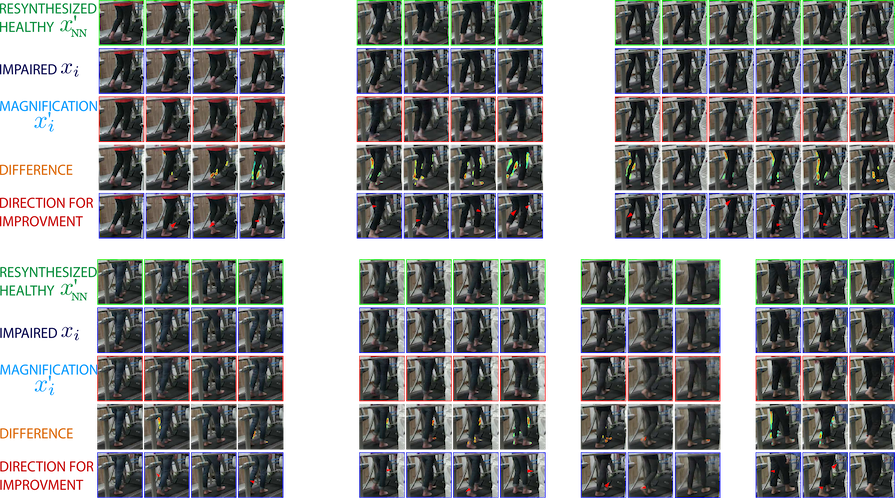}
    \caption{\textbf{Additional results of our behavior magnification for the human dataset.} See Figure 6 from the main paper for more details.}
    \label{fig:quantComp3}
\end{figure}

\clearpage
\subsection{Behavior Magnification}
In this section we provide additional qualitative and quantitative evaluations of our behavior magnification and compare our results with the latest motion magnification approach\cite{deepmag} by Oh et al. using their published implementation on github \footnote{\url{https://github.com/12dmodel/deep_motion_mag} (accessed August 2019)}.
\\ \\
In addition to Fig. 6 of the main submission, we present further results in Supplementary Figure 2 and Extended Data Figure 2. The latter figure also provides a direct comparison to the approach of Oh et al.\cite{deepmag}. We see, that our approach (4th row) successfully magnifies the discrepancies between healthy (1st row) and impaired (2nd row), whereas the motion magnification approach by Oh et al.\cite{deepmag} (3rd row) is in most of the cases not able to amplify the differences accordingly. The method by Oh et al. is not suited for analyzing posture differences across subjects (which is essential for behavior magnification). Therefore, it results in rather blurry and unrealistic looking images without magnifying the differences in the walking cycle. To help the reader spot subtle deviations, neuroscientists had afterwards added magenta makers.
\\ \\
Extended Data Figure 3 provides an additional detailed analysis of our magnification. In particular it proves that our approach only magnifies the deviations due to impaired behavior and is not fooled by the natural variations between healthy subjects. For that, we compute the intensity differences (L2 distance in the pixel space) between an original frame and its magnification. To guarantee that any differences we will observe are only due to the magnification, the original frame is also processed, but with a magnification factor of $\lambda=0$ (see Method in main manuscript).
The differences between the unmagnified original frame and its magnification indicate per pixel if there is deviation (dark) or not (white), thus localizing impairment.
Extended Data Figure 3a shows several outputs using our magnification approach (second row) and that of Oh et al. (third row). Evidently our approach is not misled by natural deviations that occur also between healthy subjects. Our magnification only highlights disease specific deviations that results from abnormal change in posture, but not natural differences that occur among healthy subjects.
In contrast, the differences between the original and the magnifications generated by Oh et al. show that their approach is not able to focus on disease specific symptoms of the subject and therefore magnifies also natural variations between healthy subjects or those occurring in the background.
To confirm the visual conclusions described above, we furthermore perform a quantitative evaluation based on the previous intensity differences. We push to zero the deviation below a certain threshold to reduce the clutter, and average the deviation for each subject and video over a time period of around $7$ seconds (200 frames). The average are then normalized onto the unit range $[0,1]$ to provide a score that indicates the degree of deviation of each of these sub-sequences. To compare the measure deviations of all healthy and all impaired subjects, Extended Data Figure 3b displays the distributions over these scores for our method (solid lines) and the approach by Oh et al. (dashed lines) using kernel density estimation with Gaussian kernel. 
Extended Data Figure 3c shows the means and standard deviations of these distributions of deviation scores. Both quantitative plots show that our approach does not discover a significant number of abnormal deviations for the healthy subjects. In contrast, for patients with impairment, the measured degree of deviation is higher and has a large range. Since impairment typically affects different parts of a gait cycle differently, e.g. one leg can be more affected than the other, we would expect such a wide range of different deviations. The approach by Oh et al. on the other hand shows a similar degree of deviation for healthy and impaired subjects, leaving them indistinguishable. The p-values of a two-sided t-test confirm this outcome. We test the null-hypothesis whether the distributions of healthy and impaired deviation scores are actually the same distribution. A p-value smaller than the significance level of $0.05$ leads to a rejection of the null-hypothesis. Our approach has a p-value of $2\times 10^{-8}$, indicating two different distributions, while the approach by Oh et al. has a p-value of $0.13$, thus not passing the significance test. Furthermore, using a Naive Bayes classifier trained on healthy and impaired deviation scores, we obtain a test accuracy of $92\%$ with our approach, and $54\%$ (chance level) with the method by Oh et al. This indicates that based on our magnification a direct separation between healthy and impaired behaviour is possible in contrast to the method of Oh et al.

\begin{table}[t]
\centering
\begin{tabular}{|l|cc|}
\hline
\textbf{Model} & \textbf{Accuracy(\%)} & \textbf{P-value}\\
\hhline{|-|--|}
Oh et al. \cite{deepmag}  & $54\%$ & $0.13$ \\
Our & \textbf{$92\%$} & \textbf{$2\times 10^{-8}$} \\
\hline  
\end{tabular}
\caption{Classifying healthy and impaired subjects using magnified postures. A p-value of $2\times 10^{-8}$ and an accuracy of $92\%$ indicate that our approach properly magnifies in a meaningful manner supporting a correct separation of healthy from impaired, in contrast to Oh et al.\cite{deepmag} that achieve no significant accuracy.}
\label{tab:ohetal}
\end{table}

\subsection{Comparison with a Standard Unsupervised Method}
To evaluate our unsupervised posture representation, we compare against a classical baseline: Principal Component Analysis (PCA). In particular, we contrast our representation against PCA on the task of posture similarity.\\
Extended Data Figure 4a (left) shows that PCA is able to identify the gait cycle consistently during the video. The same is true for our representation (right), additionally it produces smoother curves. However, when trying to relate postures from different subjects, one of our main task, PCA tends to separate videos by appearance. Even after subtracting the appearance bias, by subtracting the video average and centering each video, inter-subject differences in appearance still dominate PCA, thus hindering a comparison of posture across subjects. This, however, would be crucial for behavior analysis, therefore PCA is not a suitable approach.\\
In contrast, our representation overcomes the subject bias and pushes similar postures close to each other, independently from the appearance.\\
Extended Data Figure 4 shows only 10 subjects to keep the plot readable, but the same result is obtained when using all subjects. The original frames are shown in the attached files:\\ 6\_Unsupervised\_Method\_Comparison.

\subsection{Alternative Disentanglement Approach}
In Extended Data Figure 5, we compare our disentanglement of posture and appearance with a simple baseline. The goal of this experiment is to show that disentanglement is not an easy task and cannot be solved with simple methods. On the other hand, existing methods need fully supervised posture annotations\cite{vunet}. \\
In the baseline, appearance is extracted by averaging the frames across each video. The posture is computed by subtracting the appearance in the pixel domain from each frame. Finally, a new image is generated using the appearance from a video and the posture from a different video.\\
The results show that the baseline fails to transfer the appearance and it looses crucial posture details. On the other hand, our approach produces realistic and detailed images.

\subsection{Studying the Training Size for DeepLabCut}
In the main manuscript, we compare our unsupervised behavior representation with DeepLabCut (DLC). In this section, we study the performance of this supervised detection approach relative to the training-set size. Extended Data Figure 6 shows that a growth of $\times10$ samples improves the result by only $20\%$. To reach the performance of our unsupervised model, DLC would need a $66\%$ gain over the best observed DLC result, hence requiring around $~7000$ training samples, or $\sim100K$ clicks (i.e. $14$ keypoints per sample) by a human annotator.

\subsection{Behavior Analysis with Alternative Representation}
In addition to previous methods for behavior analysis (DeepLabCut and Jaaba), we compare our representation with modern video analysis networks. In particular, we use R3D\cite{r3d} trained on Kinetics400 as implemented by PyTorch (Extended Figure 7). 
Differently from our architecture, which uses a combination of a 2D CNN backbone for posture and an LSTM for temporal encoding, R3D is composed of 3D convolutions where each filter runs in space and time. Therefore, it does not need a dedicated temporal encoding since the two encodings (space and time) are learned jointly within the same filters. Notice that our 2D backbone gives the model the ability to explicitly represent posture information. Using 3D convolutions lacks those crucial features since we do not have control over the mixture of spatial and temporal information. Nevertheless, R3D is a good choice only the behavior encoding without posture.
Due to their similar setup, we can evaluate our model and R3D using the same protocol. \\
The results show that, even if R3D has been trained on $~300K$ labelled videos of human activities, our representation is clearly better suited for behavior analysis. Interestingly, R3D performs as good as, if not better, than DLC, showing the potential of non-parametric models.

\subsection{Detecting Key-points from Our Features}
To what extend can we recover a parametric model (key-points) from our non-parametric representation? 
The following experiment investigates how much information related to keypoint localization has been captured by our own model although its training focused on a different task.
In this experiment we regress the key-point location given the posture representation $E_\pi$. We use a two-layer network (256-128 nodes each) which receives our representation $E_\pi$ in input and outputs the 14 keypoints, with a total of 28 outputs considering x and y coordinates. We train the model on 900 frames and tested on 350. Train and test subjects are not overlapping. The performance is evaluated using Euclidean distance measured in pixels.\\
Our model is compared to a non-parametric and a parametric method: AlexNet pre-trained on ImageNet and DeepLabCut (DLC). For the former we use the same evaluation protocol as our model.
The latter is trained end-to-end since the output is already the 2D key-point location. DLC is pre-trained on MPII Human Pose Dataset\cite{mpii} and fine-tuned on our task as instructed in the DLC paper\cite{deeplabcut}.\\
As shown in Supplementary Table 6, Imagenet has an error of $19.0$ pixels and DLC reaches $10.8$ pixels.
The performances of our model is intermediate, with an error of $15.6$ pixels. These results show that our posture representation trained without human annotation is better than a model trained on more than a million labelled samples. 
However, as it was to be expected, DLC achieves a higher performance than our approach on this specific task: First, DLC has been trained specifically for predicting keypoints, whereas our approach has learned a representation for a different task and we here only implicitly recover keypoints to the extend that they were learned to be useful for the behavior analysis that our model was optimized for. Second, we have here evaluated keypoint displacement using Euclidean distance, which is also how DLC has been trained. However, as the evaluation in Supplementary Figure 3 indicates, this distance is not good measure of posture.

Even significant posture deviations can yield to low Euclidean keypoint distance with similar postures even producing higher distances. This is due to discrepancies in scale and translation between query and reference. DLC is strongly biased towards them, since it was trained on Euclidean distance. Our method is invariant to these properties, as they do not matter for behavior analysis, and focuses on the posture itself.
Extended Data Figure 8 shows qualitative results for our approach and DLC. 

\begin{table}[t!]
    \centering
    \begin{tabular}{l|cc}
     Features & Unsupervised & Error (px) \\ 
     \hline
     Imagenet & $\times$ & 19.0 \\
     DLC & $\times$ & 10.8 \\
     Ours & \checkmark & 15.6
    \end{tabular}
    \caption{\textbf{Key-points detection}. The error is measured in pixels. We train on 900 samples and test on 350. Different subjects are used for train and test. DeepLabCut (DLC) outputs the keypoints directly. For the Imagenet pre-trained AlexNet and our model, we extract the representation per each frame in the train and test set, then train a shallow network to regress the key-point from the non-parametric representation. Our representation has been trained without using labels, i.e. unsupervised.}
    \label{tab:my_label}
\end{table}

The reason why we are worse than DLC at detecting keypoints, even though we are better at classifying behavior, resides in the Euclidean distance used to train DLC and typically used to evaluate keypoint detection. However, this distance is not a good measure of posture, as it is shown in Supplementary Figure \ref{fig:kpqualitative}. Even when the Euclidean distance is low, the posture can still differ between a query and reference samples, while a reference sample with higher distance could actually have a similar posture. This is due to discrepancies in scale and translation between query and reference. Thus, DLC is strongly biased towards them since it was trained on Euclidean distance. Our method is invariant to these properties and focuses on the posture itself.

\begin{figure}[t!]
    \centering
    \includegraphics[width=0.55\linewidth]{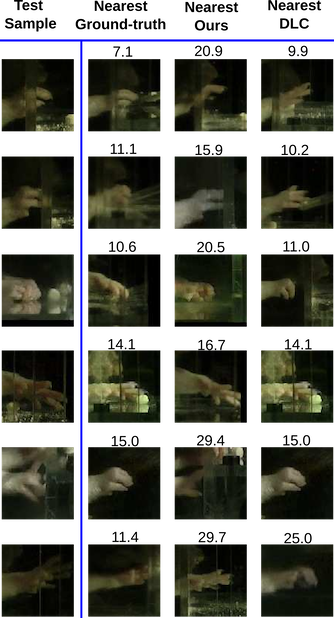}
    \caption{\textbf{Euclidean distance is wrong for posture.}. Given a test sample, we show the nearest neighbor from the train samples with the smallest posture displacement. The displacement is measured in three different way: using Euclidean distance on the manually annotated keypoints (Nearest ground-truth), using similarity based on our posture features (Ours) and Euclidean distance on the keypoints inferred by DeepLabCut (DLC). The number on top of each nearest neighbor is the Euclidean distance to the test sample based on ground-truth keypoints. Notice how our postures better reflects the test sample even though the Euclidean distance on keypoints is much higher than DLC. For example, in the first row the query paw is closed while the nearest paw using DLC shows extended fingers. On the other hand, using our representation retrieves a similar posture. The keypoints distance does not reflect this reality since DLC has an error of only $9.9$ while our method has $20.9$.}
    \label{fig:kpqualitative}
\end{figure}

\clearpage
\subsection{Quantitative Analysis of Impairment Magnification}
Our previous experiments show how behavior magnification can be beneficial for the user to amplify impaired motion of a patient. In this section, we provide a quantitative analysis showing that our magnification facilitates the identification of deviating postures without introducing artifacts that would entail misinterpretations. In particular, we set up this experiment as an anomaly detection problem on the human gait data: given train data of only normal postures (healthy), detect deviating postures within full video recordings. The dataset consists of two cohorts, healthy and impaired. For each original frame we synthesize the magnified frames following the procedure described in "Magnification process" on page 18 of the main manuscript. For each video, we limit the number of frames ($200$ or $\sim7$ full cycles) to have a uniform distribution of frames per video. \\
Now we can evaluate whether impairment is easier to detect after our magnification than in the original footage and whether this leads to misinterpretations. We extract Resnet18 features pre-trained on ImageNet (from PyTorch-Vision) and on top apply a one-class classifier, a standard one-class linear svm (from scikit-learn), to discover abnormal postures. This is a single classifier trained only on original frames of healthy patients. The classifier is then applied to recordings of different healthy and impaired subjects, original and after magnification, to predict if postures are abnormal or not. Extended Data Figure 9 shows that for healthy subjects magnification does not alter the amount of detected abnormalities ($\sim2\%$), even though the one-class classifier has been trained only on the original data and not on the magnification. This confirms that our magnification does not artificially alter correct postures. On the other hand, in the impaired cohort, our magnification has doubled the amount of detected abnormal postures, from $5.7\%$ in the original frames to $11.7\%$ after magnification (note that obviously not all video frames of impaired subjects show abnormal postures).

\clearpage
\subsection{Qualitative Analysis of Optogenetics}
In this evaluation we present two typical sequences from the same animal (Treatment cohort), one for optogenetic inhibition (light-on) and the other without (light-off). We apply the binary classifier from Fig.5b to produce a score for the two sequences. The classifier correctly assigns a negative score for "light-off" and a positive for "light-on". \\
In Extended Data Figure 10 we show five linearly-spaced frames from each sequence. We show the full frame to provide more context for the reader, including the fiber optics on the skull. However, the classifier can obviously not see the light in the glas fiber since the input image is cropped so it does not include this area. Moreover, the figure shows the ground-truth label ("GT") and classifier score ("SVM-Score") on top of each sequence. \\
The light-on sequence shows a typical failed grasp. The same animal performs a successful grasp when the laser is off. Even though the two sequences are very similar, our model successfully detects the status of the optogenetic inhibition by looking at the behavior.

\clearpage
\section{VIDEOS AND CODE}
Apart of supplementary figures, we provide additional material in the form of videos and source code files. Most of the material is also available on our \href{http://utabuechler.github.io/behaviorAnalysis}{webpage}\footnote{http://utabuechler.github.io/behaviorAnalysis}.

The material is grouped into the following folders:
 0\_code, 1\_interface, 2\_NN, 3\_VideoAnalysis, 4\_disentanglement, 5\_magnification, 6\_Unsupervised\_Method\_Comparison.
The content of these folders will be discussed one-by-one in the successive paragraphs.

\subsection{0\_code} This folder contains the code and the resources to run the interface. The scope of the interface is to test by hand the representation learned using our unsupervised method and the proposed behavior magnification.  \textit{The full code of the framework \textit{uBAM}, including also the training procedure can be found on \href{https://github.com/utabuechler/uBAM}{github}}. 


To run the code locally, all necessary resources are provided in the \textit{0\_code} folder. However, you will need to setup the anaconda environment locally on your own and install all the necessary libraries. In particular, the folder contains:

\textit{behaviorAnalysis/:} includes the python code for the interface.

\textit{bin/:} provides the commands and arguments to run the interface.

\textit{resources/:} holds a subset of the data and trained models for the human and rat dataset. The folder can be downloaded at \\ \url{https://heibox.uni-heidelberg.de/d/c67db8fa39474c13a5c9/}\\
(The files are too large to be directly included in the supplementary material.)

\subsection{1\_interface} This folder includes video tutorials explaining the functionality of the interface and how to use it. The videos are also available on our \href{http://utabuechler.github.io/behaviorAnalysis}{webpage}.

\subsection{2\_NN} Here, we provide more qualitative results of our learned encoding by showing posture and sequence Nearest Neighbors, extending the results shown in Fig. 2a and 2b of the main manuscript. The queries are chosen randomly. For the posture Nearest Neighbors (\textit{2\_NN/postures}), the *.jpg files contain the query postures in the first column and their 10 Nearest Neighbors in column 2 to 11, followed by the average of the 100 Nearest Neighbors in the last column. The files contained in \textit{2\_NN/sequences} show the query sequence in the first row and its Nearest Neighbors in the succeeding rows.

\subsection{3\_VideoAnalysis} One goal of our approach is to recognize healthy and impaired motion solely based on unprocessed video frames.
This folder contains two examples, one for a treated and the other for an untreated animal.
Based on our behavior representation, we can study the animal over the video and recognize weather the grasping function shows healthy or impaired characteristics. In case the animal has mostly healthy behavior, we visualize the most healthy-looking grasps and the three nearest sequences from the healthy references. If the behavior is mostly impaired, we visualize the most impaired sequences and provide three nearest sequences from the impaired references.
\\
The video of the treated animal shows that the sequences are mostly classified as healthy. The healthiness score is at the highest when the animal closes the paw when grasping the sugar, suggesting that the model learned that a firm grasp is characteristic of healthy animals. For the not-treated animal the model does not recognize healthy sequences since the animal does not properly close the paw (second 5, 10 and 14).

\subsection{4\_disentanglement} Our framework includes a novel unsupervised generative model for magnification which produces images by separating posture and appearance information. A qualitative evaluation of this model is given in Fig. 2c, showing generated images using posture and appearance from different subjects. The folder provides more examples, extending the evaluation of Fig. 2c to full gait sequences in the form of videos. In particular, the appearance of the query subject (columns) is transferred to the walking subject (rows). This evaluation shows that our model successfully separates the posture information from appearance. Moreover, it can generate new real-looking images, crucial for magnification.

\subsection{5\_magnification} 
In this folder, we provide video clips of the magnification differences shown in Supplementary Fig. 5 and a video demonstrating the magnification result of several walking sequences.
\\
In the folder \textit{comparison}, we provide clips in \textit{gif} format of the difference shown in Supplementary Fig. 5. For all healthy subjects, our magnified generations differ only slightly from the original behavior indicated by the almost pervasive white clips (white = no difference). However, for impaired subjects, we can see an intensity change for specific postures indicating a substantial deviation between the original and magnified frame. Moreover, notice that no difference appears in the background, while the deviation is found only in the posture. This shows once again that our posture representation has extracted the posture information and disregards the appearance. For comparison, we performed the same experiment using modern approaches for video motion magnification \cite{deepmag}. The resulting clips are very noisy and show that the approach is not able to ignore irrelevant variations between healthy subjects or deviations in the background.
\\
The video (\textit{magnification$\_$humans$\_$tsne}) provided in this folder combines the learned posture representation and magnification process to find when and where the behavior of an unhealthy subjects deviates from healthy.
\\ \\
\textit{Low Dimensionality Embedding (right):} This plot shows trajectories of two subjects using our posture representation. In these trajectories it is possible to spot when the two gaits deviate the most.
\\ \\
\textit{Magnification (bottom-center):} Using our generative model, we can amplify the behavioral difference between healthy and query subjects to facilitate the diagnosis. In this example, the query subject performs shorter steps in respect to an healthy reference subject. This difference in behavior is not directly visible in the original video, but it becomes apparent in the magnified video.
\\ \\
\textit{Difference (bottom-left):} This sequence shows the difference within the image between query and magnified sequence, suggesting which body-part mostly influences the deviation in behavior. The result shows that the impairment is focused mostly on the feet location in respect to other aspects of the motion, like leg postures. Moreover, the highest difference matches in time with the deviation between the two trajectories in the posture space (right).
\\ \\
\subsection*{6\_Unsupervised\_Method\_Comparison.}
In this folder, we provide high resolution images of Extended Data Figure 4 (c and d) showing the original RGB frame instead of the scatter plot.

\clearpage
\section*{REFERENCES}
\vspace{15px}
\bibliography{ms}{}
\bibliographystyle{naturemag}